\documentclass{article}

\usepackage{microtype}
\usepackage{graphicx}
\usepackage{subfig}
\usepackage{booktabs} 
\usepackage{amsmath}
\usepackage{hyperref}

\usepackage[accepted]{icml2025}
\usepackage{amsmath}
\usepackage{amssymb}
\usepackage{mathtools}
\usepackage{amsthm}
\usepackage{multirow}
\usepackage{algorithm}
\usepackage{algpseudocode}
\usepackage{amsmath}

\usepackage{colortbl}  
\usepackage{xcolor}
\usepackage{array}   

\definecolor{lightyellow}{RGB}{255,248,220}
\definecolor{lightblue}{RGB}{220,235,255}
\definecolor{lightgray}{RGB}{245,245,245}
\usepackage[capitalize,noabbrev]{cleveref}

\theoremstyle{plain}

\theoremstyle{definition}

\theoremstyle{remark}

\usepackage[textsize=tiny]{todonotes}

\icmltitlerunning{SMV-EAR: Bring Spatiotemporal Multi-View Representation Learning into Efficient Event-Based Action Recognition}

\begin{document}

\twocolumn[
\icmltitle{SMV-EAR: Bring Spatiotemporal Multi-View Representation Learning into Efficient Event-Based Action Recognition}



\begin{icmlauthorlist}
\icmlauthor{Rui Fan}{}
\icmlauthor{Weidong Hao}{}

\end{icmlauthorlist}

\vskip 0.3in
]




\begin{abstract}
Event cameras action recognition (EAR) offers compelling privacy-protecting and efficiency advantages, where temporal motion dynamics is of great importance. Existing spatiotemporal multi-view representation learning (SMVRL) methods for event-based object recognition (EOR) offer promising solutions by projecting $H$-$W$-$T$ events along spatial axis $H$ and $W$, yet are limited by its translation-variant spatial binning representation and naive early concatenation fusion architecture. This paper reexamines the key SMVRL design stages for EAR and propose: (i) a principled spatiotemporal multi-view representation through translation-invariant dense conversion of sparse events, (ii) a dual-branch, dynamic fusion architecture that models sample-wise complementarity between motion features from different views, and (iii) a bio-inspired temporal warping augmentation that mimics speed variability of real-world human actions. On three challenging EAR datasets of HARDVS, DailyDVS-200 and THU-EACT-50-CHL, we show +7.0\%, +10.7\%, and +10.2\% Top-1 accuracy gains over  existing SMVRL EOR method with surprising 30.1\% reduced parameters and 35.7\% lower computations, establishing our framework as a novel and powerful EAR paradigm.
\end{abstract}

\section{Introduction}
\label{sec:1}

\begin{figure}[t]  
\centering  
\includegraphics[width=1.0\linewidth]{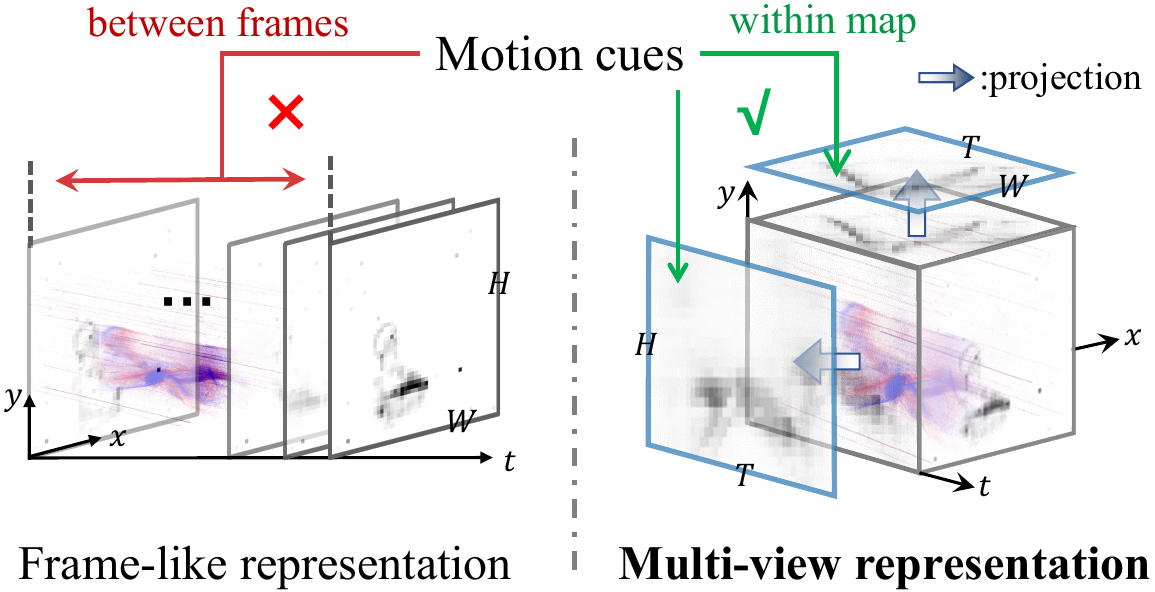}  
\caption{Comparison of frame-like representation in prior SOTA EAR method (left) and our adopted multi-view representation (right). Our SMV-EAR embeds motion cues within $T$-$H$ and $T$-$W$ maps rather than between $H$-$W$ frames.}  
\label{fig:1}  
\end{figure}

Event cameras perceive luminance changes asynchronously with high temporal resolution ($\sim$1$\mu$s), appearance-free streaming and low power consumption~\cite{gallego2020event,chakravarthi2024recent,miao2019neuromorphic,steffen2024efficient,deng2021mvf,wang2024dailydvs}. Existing state-of-the-art (SOTA) event-based action recognition (EAR) methods convert $H$-$W$-$T$\footnote{In this paper, we use H,W,T to denote the spatiotemporal resolution and $H$,$W$,$T$ to denote the spatiotemporal dimensions (or axes).}   sparse events into frame-like representations through temporal binning\footnote{Discretize the time axis ($T$) into consecutive bins~\cite{deng2021mvf}.} (Fig.~\ref{fig:1} left) then adapt well-trained 2D models~\cite{wang2024dailydvs,wang2024hardvs,xie2024event}. Although familiar to human vision, however, this practice embeds the fundamental temporal motion cues between the evolving frames over $T$ axis with limited frame count, which may fall short in fully capturing and modeling physically continuous human actions~\cite{li2019collaborative,wang2024hardvs}. 


\begin{figure}[t]  
\centering  
\includegraphics[width=0.85\linewidth]{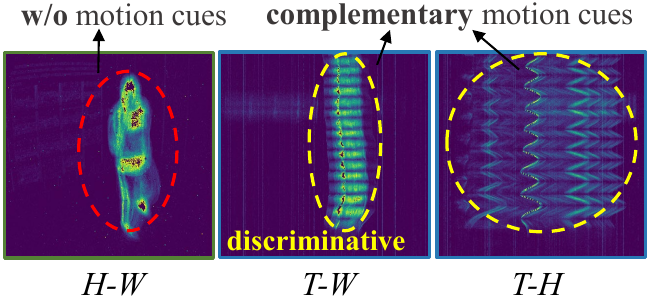}  
\caption{Projections of an action sample on different views.}  
\label{fig:example}  
\end{figure}

To tackle this, we propose a \textbf{S}patiotemporal \textbf{M}ulti-\textbf{V}iew\footnote{`Spatiotemporal multi-view' refers to multiple viewpoints of spatiotemporal data with $H$-$W$-$T$ dimenstions, e.g. video and events~\cite{li2019collaborative,berahmand2025comprehensive}.} representation learning (SMVRL) framework for \textbf{EAR} task (\textbf{SMV-EAR}). As shown in Fig.~\ref{fig:1} right, instead of aggregating events into $H$-$W$ (Height-Width) frames along $T$ axis, it projects events along $H$ and $W$ axes, which embeds temporal motion cues within the projected $T$-$W$ (Time-Width) or $T$-$H$ (Time-Height) maps rather than between the $H$-$W$ frames. While unfamiliar to human beings, these 2D maps record fine-grained and discriminative  complementary motion cues (Fig.~\ref{fig:example}), thereby exhibiting significant potential for achieving more accurate and efficient EAR.  



\begin{figure}[t]
    \centering
\includegraphics[width=0.95\linewidth]{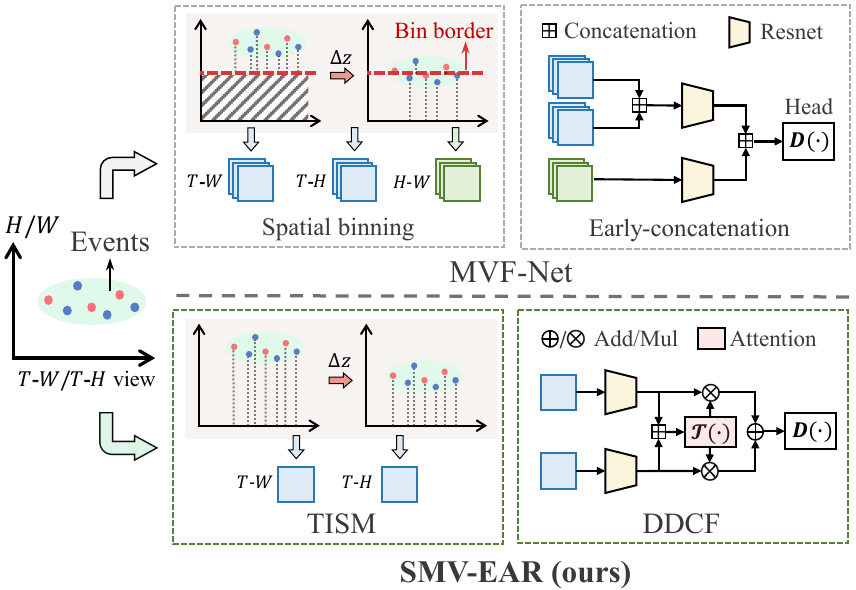}
    \caption{Comparisons of representation and architecture between baseline SMVRL method~\cite{deng2021mvf} (top) and our SMV-EAR (bottom). SMV-EAR ensures translation-invariant and reasonable SMVRL.}
    \label{fig:3}
\end{figure}


However, directly applying frame-like representations for $T$-$W$ and $T$-$W$ views will inevitably degrade the EAR performance. This stems from two key problems: (i) An action type is expected to be independent from ``where it happens'' (spatial location). However, naively adapting $T$-axis binning for the spatial projection $H$/$W$ axes (\textbf{spatial binning}) may fragment coherent action events (Fig.~\ref{fig:3} left top), which will lead to unwanted translation-variant representations thereby resulting in poor feature discriminability (Fig.~\ref{fig:5}). (ii) Naive \textbf{early concatenation} of $T$-$H$ and $T$-$W$ maps~\cite{deng2021mvf} may overlook their inherent dimension misalignment ($H$ vs. $W$) and semantic discrepancy (horizontal vs. vertical motion, Fig.~\ref{fig:example}), thereby failing in fully exploiting cross-view complementarity (Fig.~\ref{fig:3} right top). 


In this paper, we address these challenges by examining critical design stages of SMV-EAR: (i) By examining a plenty of basic conversion windows, measurements and aggregations~\cite{zubic2023chaos}, we propose a \textbf{T}ranslation-\textbf{I}nvariant \textbf{S}patiotemporal \textbf{M}ulti-view (\textbf{TISM}) representation that embeds temporal motion cues with global event count and polarity information for principled translation-invariance guarantee and compactness (Fig.~\ref{fig:3} left bottom); (ii) By examining diverse extraction and fusion architectures, we propose a \textbf{D}ual-branch \textbf{D}ynamic \textbf{C}ross-view \textbf{F}usion (\textbf{DDCF}) architecture that extracts view-specific action logits independently and dynamically re-weights them with cross-view attention derived from sample-wise semantics (Fig.~\ref{fig:3} right bottom); (iii) Consider the diverse speed variance of real-world human actions~\cite{feichtenhofer2019slowfast}, we additionally establish \textbf{D}iverse \textbf{T}emporal \textbf{W}arping (\textbf{DTW}) as a critical EAR augmentation that generates continuous, view-consistent action variations to further improve performance (Fig.~\ref{fig:7}).

We show on three existing most challenged EAR benchmarks, including HARDVS~\cite{wang2024hardvs}, DailyDVS-200~\cite{wang2024dailydvs} and THU-EACT-50-CHL~\cite{gao2023action}, that our SMV-EAR achieves +7.0\%, +10.7\%, and +10.2\% Top-1 accuracy gains respectively, while dramatically cutting parameters by 30.1\% and computations by 35.7\% compared to the base line SMVRL method, MVF-Net~\cite{deng2021mvf} (Fig.~\ref{fig:comparison}). This establishes our SMV-EAR as a novel and powerful SMVRL EAR paradigm. Our main contributions can be organized as:

\begin{figure}[t]  
\centering  
\includegraphics[width=0.90\linewidth]{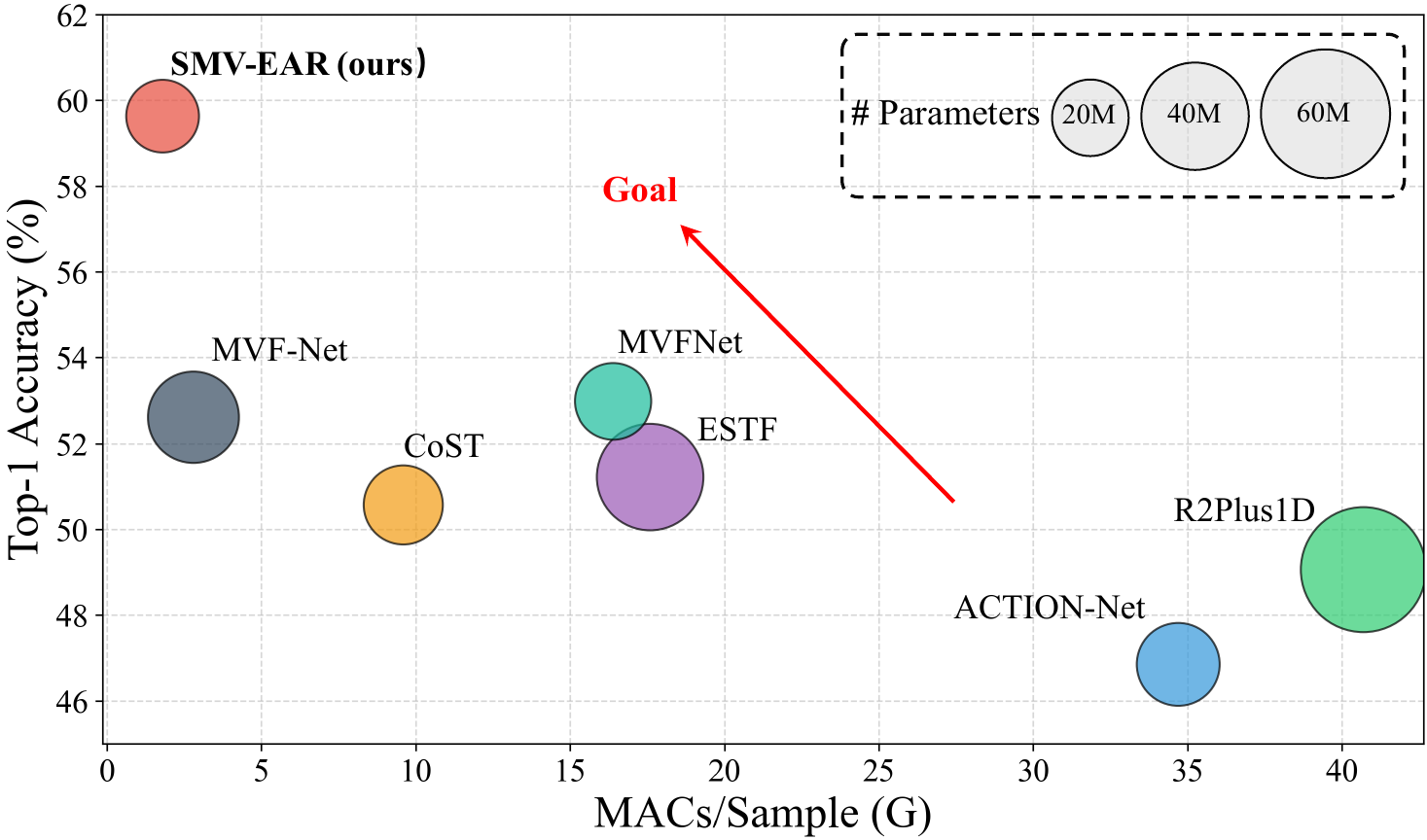}  
\caption{Results on HARDVS dataset. Our SMV-EAR surpasses all methods and sets a new performance frontier for EAR task.}  
\label{fig:comparison}  
\end{figure}  
 

\begin{itemize}
\item While frame-like representation on H-W frame is friendly for human perception, we argue that projection on Time-$W$,$H$ views can better capture temporal motion for EAR task. We also identify two key bottlenecks, spatial binning and early concatenation for this presentation. 
\item We propose a SMVRL framework for EAR task, which at first uses a TISM representation to encode fine-grained temporal motion cues within maps while enabling translation-invariance. Then, a DDCF architecture is designed to exploit sample-wise complementarity between heterogeneous temporal views. Finally, we introduce the DTW, a simple but fundamental augmentation to mimic speed variability of real-world human actions.
\item On challenging EAR benchmarks, our method consistently achieves significant accuracy gains over the prior SOTA methods with roughly 70\% computations.
\end{itemize}


\section{Related Work}
\label{sec:2}
\subsection{Event-Based Action Recognition} 
Existing SOTA EAR methods convert sparse events into frame-like representations~\cite{rebecq2017real,zhu2019unsupervised,gehrig2019end} to process them with well-trained CNNs~\cite{he2016deep} or Video Transformers~\cite{neimark2021video,liu2022video}. Despite superior in accuracy, aggregating events over discrete, limited time bins inevitably loses fine-grained temporal dynamics~\cite{schaefer2022aegnn}. Alternative point-based methods treat events as spikes~\cite{lin2024spike,ren2023spikepoint,chen2024spikmamba}, point clouds~\cite{sun2025event}, or graphs~\cite{xie2022vmv,sun2023asynchronous,deng2024dynamic} to employ architectures like SNNs~\cite{wu2018spatio,tavanaei2019deep}, point networks~\cite{qi2017pointnet,qi2017pointnet++}, or GCNs~\cite{fey2018splinecnn,wang2019dynamic}. While efficient, these works fall short in accuracy and general hardware support~\cite{ren2023spikepoint,fan2025eventpillars}. There are also works aiming to address EAR challenges under unique scenarios, such as language-guided~\cite{zhou2024exact}, multi-(camera)view~\cite{gao2024hypergraph}, wildlife~\cite{hamann2025fourier}, multi-modal~\cite{steffen2024efficient,wang2025rgb} and few-shot~\cite{ruan2025few} settings, yet our focus is on general human action recognition (HAR). Although beyond the scope, our SMV-EAR is orthogonal to these works thus can be integrated into them for improved performance.

\subsection{SMVRL} 
As discussed, SMVRL offers a promising paradigm for EAR by modeling temporal dynamics from $T$-$H$ and $T$-$W$ views. However, naively adapting SMVRL methods from video analysis~\cite{li2019collaborative,wu2021mvfnet,wu2025transformer} or EOR task~\cite{deng2021mvf} proves suboptimal. This is attributed to: (i) The spatial binning requirement of theses methods causes an unwanted dependence between spatial location (on projection axis) of action events and the projected view maps; (ii) Their early concatenation~\cite{deng2021mvf} or mid-level fusion~\cite{li2019collaborative,wu2021mvfnet} prematurely mixes features from different views characterized by distinct action semantics. Our SMV-EAR features translation-invariant representation and dynamic late-fusion architecture, which respects the semantic discrepancy of different views while exploiting their complementarity effectively. 

\subsection{Data Augmentation for Event Data} 
Existing event-based augmentations include random drops~\cite{gu2021eventdrop}, spatial transformations~\cite{li2022neuromorphic}, occlusions~\cite{bendig2024shapeaug}, meta learning~\cite{gu2024eventaugment}, and mixing strategies~\cite{shen2023eventmix,dong2025eventzoom}. However, they often fail to generate complex, non-uniform temporal dynamics characterizing real-world human actions~\cite{feichtenhofer2019slowfast}. Our DTW distinguishes itself by comprehensive temporal warping functions. Moreover, video-based temporal augmentations (e.g., temporal sampling~\cite{ramesh2023trandaugment}, temporal jittering~\cite{chen2025temporal}) operate on dense frame intervals, while our DTW directly manipulates sparse event timestamps rather than frame indices to apply non-uniform warping and ensure consistent warping across different views. This is a geometric constraint unattainable in frame-based methods.

\section{Methodology} 
\label{sec:3}
\subsection{Overall Framework} 
\label{sec:3_1}
\begin{figure}[!t]
    \centering
    \includegraphics[width=0.99\linewidth]{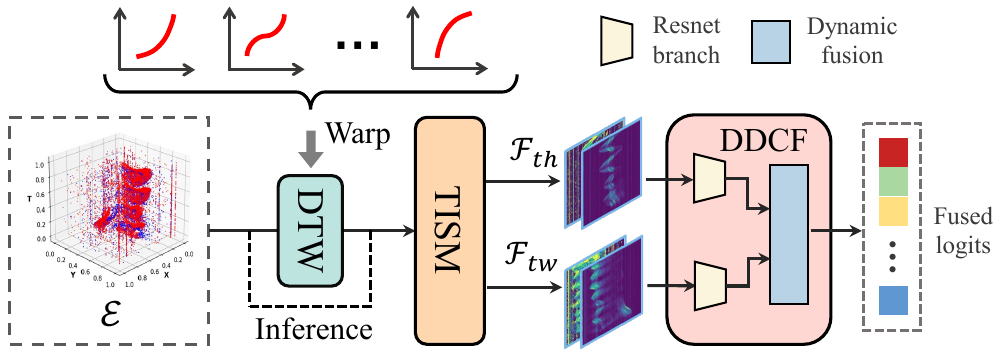}
    \caption{Overview of main contributions in SMV-EAR's pipeline.}
    \label{fig:4}
\end{figure}

Our work is built upon the principle that effective SMVRL EAR framework requires: (i) translation-invariant multi-view representation (by TISM), (ii) view-specific motion extraction and dynamic cross-view fusion (by DDCF), and (iii) diverse action variant generation (by DTW). As shown in Fig.~\ref{fig:4}, input events $\mathcal{E}$ are converted by TISM into 2D feature maps $\{F_{th}, F_{tw}\}$ from $T$-$H$ and $T$-$W$ views. DDCF processes these maps with independent neural branches and cross-view dynamic attention fusion to predict action types. Finally, DTW augmentation further enhances test accuracy.  

\subsection{Translation-invariance of TISM}
\label{sec:3_2}

\begin{figure*}[!t]
    \centering
    \includegraphics[width=0.90\linewidth]{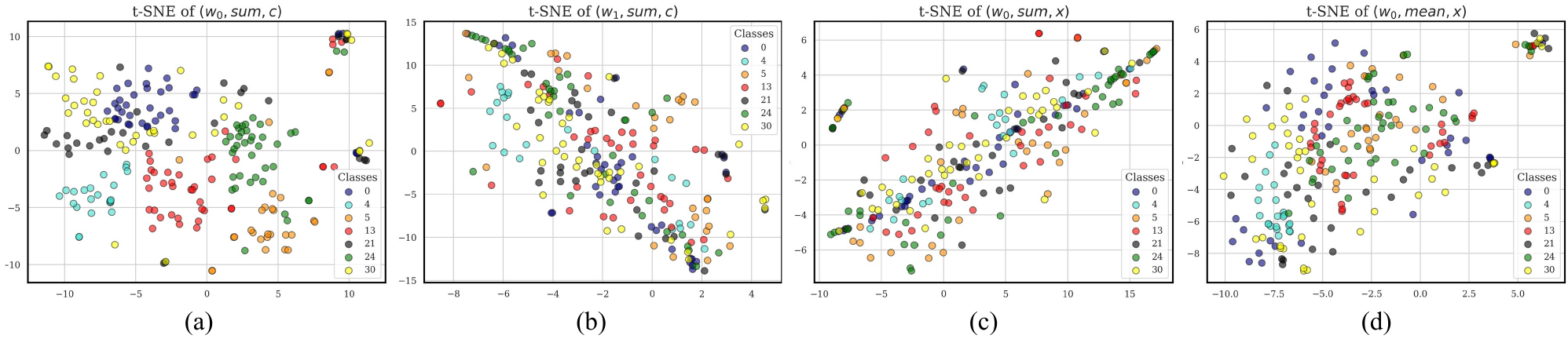}
    \caption{$t$-SNE visualization of $T$-$H$ feature maps from different encoding functions, $w_{1}$: 2-bin. (a) vs (b) shows global encoding is better than binning. (a) vs (c),(d) shows that translation-invariant measurements and aggregations is necessary for discriminative feature space.}
    \label{fig:5}
\end{figure*}

\noindent\textbf{Formulation.} Let $\mathcal{E}=\{(x_k,y_k,t_k,p_k)\}_{k=0}^{N_e-1}$ be the input events, where $(x_k,y_k)$, $t_k$, and $p_k\in\{-1, +1\}$~\cite{de2023eventtransact} are the spatial coordinate, timestamp, and polarity, respectively. For view $v\in\{th,tw,hw\}$, let $z^v$ denote the orthogonal axis ($z^{th}=x$, $z^{tw}=y$, $z^{hw}=t$). Following~\cite{zubic2023chaos}, we can decompose the encoding function into three stages:  
\begin{align}  
\label{eq:1}  
F_{v}(\mathcal{E})=[F_{0},\cdots,F_{N_c-1}], \quad F_{c} = a_c(m_c(w_c(\mathcal{E}))),  
\end{align}  
where $w_c\in\mathcal{W}=\{[t_{0,k},t_{1,k}]\}_{k=0}^{N_b-1}$ is the window function, $m_c\in\mathcal{M}=\{z^{v}_+,z^{v}_-,z^{v},p,c_+,c_-,c\}$ for measurement function, and $a_c\in\mathcal{A}=\{\textit{max},\textit{min},\textit{sum},\textit{mean},\textit{variance}\}$ the aggregation function. Hence, assuming no out-of-bounds conditions (Fig.~\ref{fig:3}), the translation-invariance means for any shift $\Delta z^v$:  
\begin{align}  
\label{eq:2}  
F_{v}(\mathcal{E}|_{z^v})=F_{v}(\mathcal{E}|_{z^v+\Delta z^v}).  
\end{align}  
This leads to find a set of conversion functions parametrized by $P=\{(w_c,a_c,m_c)\}_{c=0}^{N_c-1}$ that fulfill Eq.~\ref{eq:2}. We validate our analysis in Fig.~\ref{fig:5}, where translation-invariant $w_c$, $a_c$ and $m_c$ on temporal view map (e.g., $T$-$H$) are indeed all contributed to achieve more discriminative feature space.

\noindent\textbf{Implementation.} Through rigorous mathematical derivations (see Supplementary), we can identify a set of feasible, translation-invariant conversion functions:  
\begin{align}  
\label{eq:3}  
P_{IV}=\{&(w_0,\textit{sum},p),(w_0,\textit{sum},c),(w_0,\textit{sum},c_+), \nonumber \\
&(w_0,\textit{sum},c_-), (w_0,\textit{variance},z^v), \nonumber\\
&(w_0,\textit{variance},z^v_+), (w_0,\textit{variance},z^v_-)\},  
\end{align}  
where $w_0$ is a global, bin-less window. Through Eq.~\ref{eq:3}, the key findings include: (i) global window $w_0$ is a must; (ii) measurements $\{z^v,z^v_+,z^v_-\}$ involving $z$ dimension satisfy Eq.~\ref{eq:2} only when with $variance$ aggregation. However, since $variance$ is computationally expensive (2-order moments), we adopt 1-order $m_c=\textit{sum}$ for efficiency (Tab.~\ref{tab:5}). Moreover, since $c,p$ measurements have been proven more expressive than $c_+,c_-$~\cite{fan2025eventpillars}, we end up selecting the compact $P^{*}_{IV}=\{(w_0,\textit{sum},p),(w_0,\textit{sum},c)\}$, yielding:
\begin{align}  
F_v=[\textit{sum}(c(w_0(\mathcal{E}))),\textit{sum}(p(w_0(\mathcal{E})))]\in\mathbb{R}^{2\times U \times V},  
\end{align}  
where $U\times V$ is the 2D resolution of view $v$. This obtains translation-invariant feature while maintaining efficiency. 

\subsection{Dual-branch and Dynamic DDCF} 
\label{sec:3_3}
\begin{figure}  
    \centering  \includegraphics[width=0.90\linewidth]{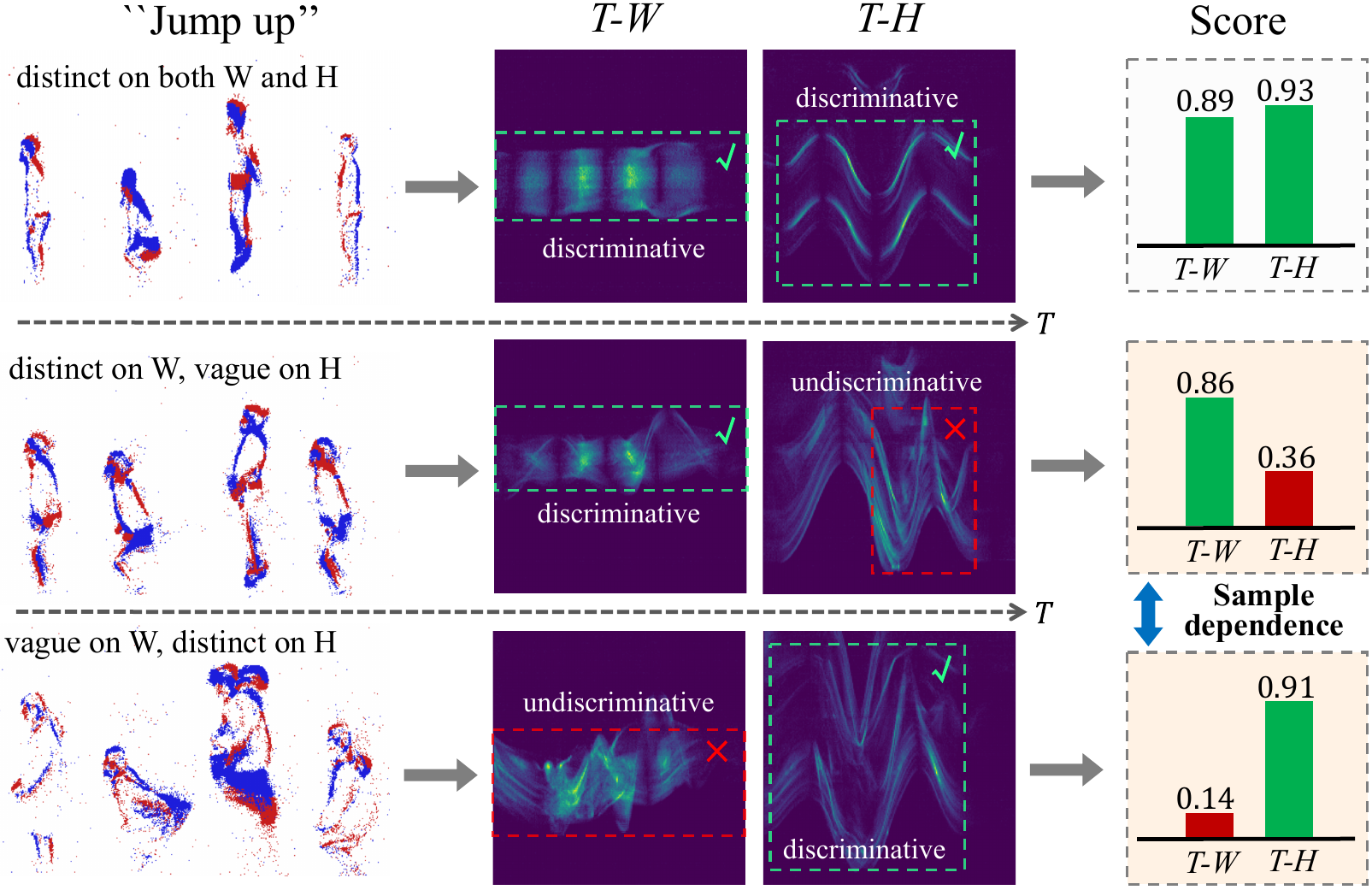}  
    \caption{Predictions of two trained ResNets on different samples under a same action class. Optimal fusion may vary with samples.}  
    \label{fig:6}  
\end{figure}

\noindent\textbf{View Selection.} Under the scheme of our global TISM representation, \citet{zhu2019unsupervised} has shown that $F_{hw}$ doesn't record EAR-critical temporal cues. This can also be witnessed in Fig.~\ref{fig:example} intuitively, where $F_{hw}$ only posses spatial context of ``where the actions arise'' (may be useful for object detection) and spatial appearance of ``what is moving'' (may be useful in tasks involving multi-kind subjects beyond humans, yet without witnessed works and out-of-focus). Hence we select $F_{th}$,$F_{tw}$ for efficiency. Ablations also show that including $F_{hw}$ increases compute/parameters with negligible accuracy improvement (Tab.~\ref{tab:7}): 

\noindent\textbf{Formulation.} Let $L = \mathcal{R}(F)$ denote a ResNet $\mathcal{R}(\cdot)$ predicting logits $L\in\mathbb{R}^C$ on $F_v$, $C$ is number of classes. MVF-Net~\cite{deng2021mvf} applies early-concatenation with shared-branch extraction $L = \mathcal{R}([F_{th}, F_{tw}])$, which ignores dimension misalignment and informational heterogeneity (Fig.~\ref{fig:example}), motivating dual-branch extraction with late fusion:
\begin{align}  
L = \mathcal{F}([L_{th}, L_{tw}]), \ L_{v} = \mathcal{R}_{v}(F_{v}),  
\end{align}  
where the fusion strategy $\mathcal{F}(\cdot)$ aims to exploit cross-view complementarity. The feasible $\mathcal{F}(\cdot)$ including logits averaging (LA), view-wise weighting (VW), and class-wise weighting (CW), with their formulations and attributes are detailed in Supplementary. However, even the same-class actions may exhibit distinct discriminability on different views, depending on its motion change and amplitude on horizontal ($W$) and vertical ($H$) dimension (e.g., a ``Jump up'' action in Fig.\ref{fig:6}). This leads us to design a sample-wise weighting $\mathcal{F}(\cdot)$ that dynamically adapt to each sample: 
\begin{align}  
L = w_{th}(\mathcal{E}) L_{th} + w_{tw}(\mathcal{E}) L_{tw}.  
\end{align}  

\noindent\textbf{Implementation} In principal, $w_{v}(\mathcal{E})$ can be learned from any feature between $F_{v}$ to $L$. However, learning from input or middle-layer features still requires to resolve dimension misalignment not to mention the large computation, while learning from $L$ may be insufficient in semantics for its high compactness. We therefore choose to learn dynamic weights from more informative, globally pooled $S_{v}\in\mathbb{R}^{512}$ before each branch's classification head that does not involve view's dimension to obtain $w_{th},w_{tw}$ (Fig.~\ref{fig:3}):
\begin{align}
\left[w_{th},w_{tw}\right] &=\mathcal{L}(\mathcal{T}(S))\in\mathbb{R}^{2C}.
\end{align}  
Here, a multi-head attention  block~\cite{vaswani2017attention} $\mathcal{T}(\cdot)$ is applied on semantic sequence $S=[S_{th},S_{tw}]$ to leverage its global modeling capability to model the cross-view complementarity. $\mathcal{L}(\cdot)$ is a linear layer. This is referred to ``dynamic'' for its consideration of sample-specific fusion weights. 

\subsection{Diverse Speed Variations of DTW}
\label{sec:3_4}
\noindent\textbf{Formulation} Real-world human actions exhibit diverse speed variations~\cite{feichtenhofer2019slowfast}, which are often neglected in existing event-based augmentations~\cite{gu2021eventdrop,li2022neuromorphic,shen2023eventmix}. For an action with $T = t_e - t_0$ duration, speed variation alters temporal event density~\cite{wu2024egsst}. Let $\mathcal{W}: \{t_k\}_{k=0}^{N_e-1}\rightarrow \{t'_k\}_{k=0}^{N_e-1}$ denote a temporal warping function mapping timestamps to warped ones, the instantaneous speed scaling factor at time $t$ is $s(t) = \frac{d\mathcal{W}(t)}{dt}$. When $s(t) > 1$, the action decelerates locally (events become sparser); when $s(t) < 1$, it accelerates (events becomes denser). Therefore, the key to mimic diverse speed variations lies in diverse, non-uniform $\mathcal{W}(\cdot)$.
 
\noindent\textbf{Implementation} We show procedure of DTW in Algo.~\ref{alg:1}. To achieve diverse $\mathcal{W}(\cdot)$, we parameterize it through multiple non-uniform functions, including \textit{identity, linear, power, exponential, cosine}. Each function introduces different speed profiles while maintaining temporal ordering (formulations and monotonicity guarantee are detailed in Supplementary), unlike disruptive augmentations such as FlipT\footnote{Flip events along time dimension as defined in~\cite{gu2024eventaugment}.}~\cite{gu2024eventaugment} that invert temporal causality (e.g., transforming ``Sit down'' into ``Stand up''). In Fig.~\ref{fig:7} a typical sample shows that our DTW yields local density modulation to match the intended speed variability. This observation effect aligns with the test accuracy gains in Sec.~\ref{sec:4_3}.

\begin{table*}[!t]
\centering
\caption{Comparisons on HARDVS and DailyDVS-200. * denotes w/o augmentation and `-' indicates unavailable results in existing publications, otherwise borrowed or replicating with their original configurations. MACs and T(all) are measured on HARDVS.}
\label{tab:2}
\scriptsize
\setlength{\tabcolsep}{2pt}
\begin{tabular}{lcc|cc|cc|cc}
\toprule
\multirow{2}{*}{\textbf{Method}} & \multirow{2}{*}{\textbf{Input Type}} & \multirow{2}{*}{\textbf{Backbone}} & \multicolumn{2}{c|}{\textbf{Acc. on HARDVS}$\uparrow$} & \multicolumn{2}{c|}{\textbf{Acc. on DailyDVS-200}$\uparrow$} & \multirow{2}{*}{\textbf{MACs}$\downarrow$} & \multirow{2}{*}{\textbf{Params}$\downarrow$} \\
& & & Top-1(\%) & Top-5(\%) & Top-1(\%) & Top-5(\%) & & \\
\midrule 
SDT~\cite{yao2023spike} & \multirow{2}{*}{Spike} & Transformer & - & -& 35.43(-) & 58.81(-) & - & 29.3M \\
Spikeformer~\cite{zhou2022spikformer} & & Transformer & - & - & 36.94(-) & 62.37(-) & - & 29.7M \\
\midrule 
C3D~\cite{tran2015learning} & \multirow{5}{*}{Frame} & 3D CNN & 50.52(-) & 56.14(-) & 21.99(-) & 45.81(-) & 0.2G & 147.2M \\
R2Plus1D~\cite{r2plus1d} & & ResNet34 & 49.06(-) & 56.43(-) & 36.06(-) & 63.67(-) & 40.7G & 63.5M \\ 
SlowFast~\cite{feichtenhofer2019slowfast} & & ResNet50 & 50.63(-) & 57.77(-) & 41.49(-) & 68.19(-) & 0.7G & 33.6M \\
TSM~\cite{lin2019tsm} & & ResNet50 & 52.63(-) & 60.56(-) & 40.87(-) & 71.46(-) & 0.7G & 24.3M \\
ACTION-Net~\cite{wang2021action} & & ResNet50 & 46.85(-) & 56.19(-) & 42.61($\pm$0.16) & 71.24($\pm$0.23) & 34.7G & 27.9M \\
\midrule
EST~\cite{gehrig2019end} & Learned & ResNet34 & 36.51($\pm$0.43) & 42.09($\pm$0.64) & 32.23(-) & 59.66(-) & 2.1G & 21.5M \\
\midrule
TimeSformer~\cite{bertasius2021space} & \multirow{4}{*}{Token} & Transformer & 50.77(-) & 58.70(-) & 44.25(-) & 74.03(-) & 107.3G & 121.2M \\
Swin-T~\cite{liu2022video} & & Transformer & 51.91(-) & 59.11(-) & 48.06(-) & 74.47(-) & 17.5G & 27.8M \\
GET~\cite{peng2023get} & & Transformer & 46.46($\pm$0.38) & 52.37($\pm$0.19) & 37.28(-) & 61.59(-) & 0.9G & 4.5M \\
ESTF~\cite{wang2024hardvs} & & ResNet18 & 51.22(-) & 57.53(-) & 24.68(-) & 50.18(-) & 17.6G & 46.1M \\
\midrule
CoST~\cite{li2019collaborative} & \multirow{5}{*}{Multi-view} & ResNet50 & 50.57($\pm$0.69) & 61.38($\pm$0.47) & 36.09($\pm$0.71) & 64.45($\pm$0.26) & 9.6G  & 25.4M \\
MVFNet~\cite{wu2021mvfnet} & & ResNet50 & 52.98($\pm$0.41) & 63.26($\pm$0.29) & 48.30($\pm$0.27) & 75.91($\pm$0.31) & 16.4G & 23.6M \\
MVF-Net~\cite{deng2021mvf} & & ResNet34\&18 & 52.61($\pm$0.39) & 61.67($\pm$0.07) & 43.98($\pm$0.18) & 70.39($\pm$0.95) & 2.8G & 33.6M \\
\rowcolor{lightblue} SMV-EAR*(ours) & & ResNet18 & 55.63($\pm$0.37) & 63.56($\pm$0.23) & 50.06($\pm$0.48) & 76.47($\pm$0.30) & 1.8G & 23.5M \\
\rowcolor{lightblue} SMV-EAR(ours) & & ResNet18 & 59.63($\pm$0.19) & 67.56($\pm$0.25) & 54.65($\pm$0.52) & 78.28($\pm$0.27) & 1.8G & 23.5M \\
\bottomrule
\end{tabular}
\end{table*}

\begin{algorithm}[!t]
\footnotesize
    \caption{Procedures of DTW augmentation.}
    \label{alg:1}
    \hspace*{0.235in}\textbf{Input:} Timestamps of event sequence $\varepsilon=\{t_k\}_{k=0}^{N-1}$.\\
    \hspace*{0.235in}\textbf{Output:} Warped timestamps $\varepsilon^*=\{t_k^{*}\}_{k=0}^{N-1}$.
    \begin{algorithmic}[1]
    \State Initialize \(\varepsilon^*\), namely \(\varepsilon^* \gets \varepsilon\);
    \State Randomly select $l$ non-overlapping intervals $\{I_j\subseteq [t_0,t_{N-1}]\}_{j=0}^{l}$;
    \For{each interval \(I_j\)} 
        \State Select $\mathcal{W}\gets$Random.choice(\textit{identity,\ linear, power, exponential,cosine});   
        \State Sample \textit{magnitude} $\{\alpha,\beta,\gamma,\eta\}\sim\mathcal{U}(a,b)$;
        \For{each timestamp \(t_k\in\varepsilon^{*}\)}
            \If{$t_k$ in $I_j$}
                \State $t^{*}_k = \mathcal{W}(t_k, \textit{magnitude})$;
                \State Replace \(t_k\) by \(t^{*}_k\);
            \EndIf
        \EndFor
    \EndFor
    \State Adjust $\varepsilon^*$ to connect the warped timestamps with original, non-warped timestamps to ensure continuity.
    \end{algorithmic}
\end{algorithm}

\section{Experiments}
\label{sec:4}
\subsection{Experimental Setup}
\label{sec:4_1}
\noindent\textbf{Datasets.} We evaluate on three recent challenging EAR datasets. \textit{THU-EACT-50-CHL}~\cite{gao2023action} provides 2,330 samples across 50 action types captured under various lighting conditions at 346$\times$260 resolution. \textit{HARDVS}~\cite{wang2024hardvs} includes 107,646 event sequences spanning 300 action types from 5 human subjects, recorded with DAVIS346 camera at 346$\times$260 resolution. \textit{DailyDVS-200}~\cite{wang2024dailydvs} consists of 22,046 sequences covering 200 daily action types from 47 human subjects, incorporating 14 distinct challenging attributes including camera motion, lighting variations, and etc..

\noindent\textbf{Evaluation Metrics.} With mean$\pm$standard deviation over five random seeds, we report Top-1 and Top-5 accuracy and measure model parameters (Params), multiply-accumulate operations (MACs), the number of floating-point operations (FLOPs) for comparison following established protocols~\cite{gao2023action,wang2024hardvs}. The inference time T(all) = T(cpu)+T(gpu) is measured with batch size 1, where T(cpu) is the sparse-to-dense events conversion time on an AMD EPYC 7542 CPU and T(gpu) the network feedforward time on a RTX 3090 GPU. For translation-robustness evaluation, we inject test-time $z$-axis translations to events and clip into H/W range. 

\begin{figure}[t]  
\centering  
\includegraphics[width=0.85\linewidth]{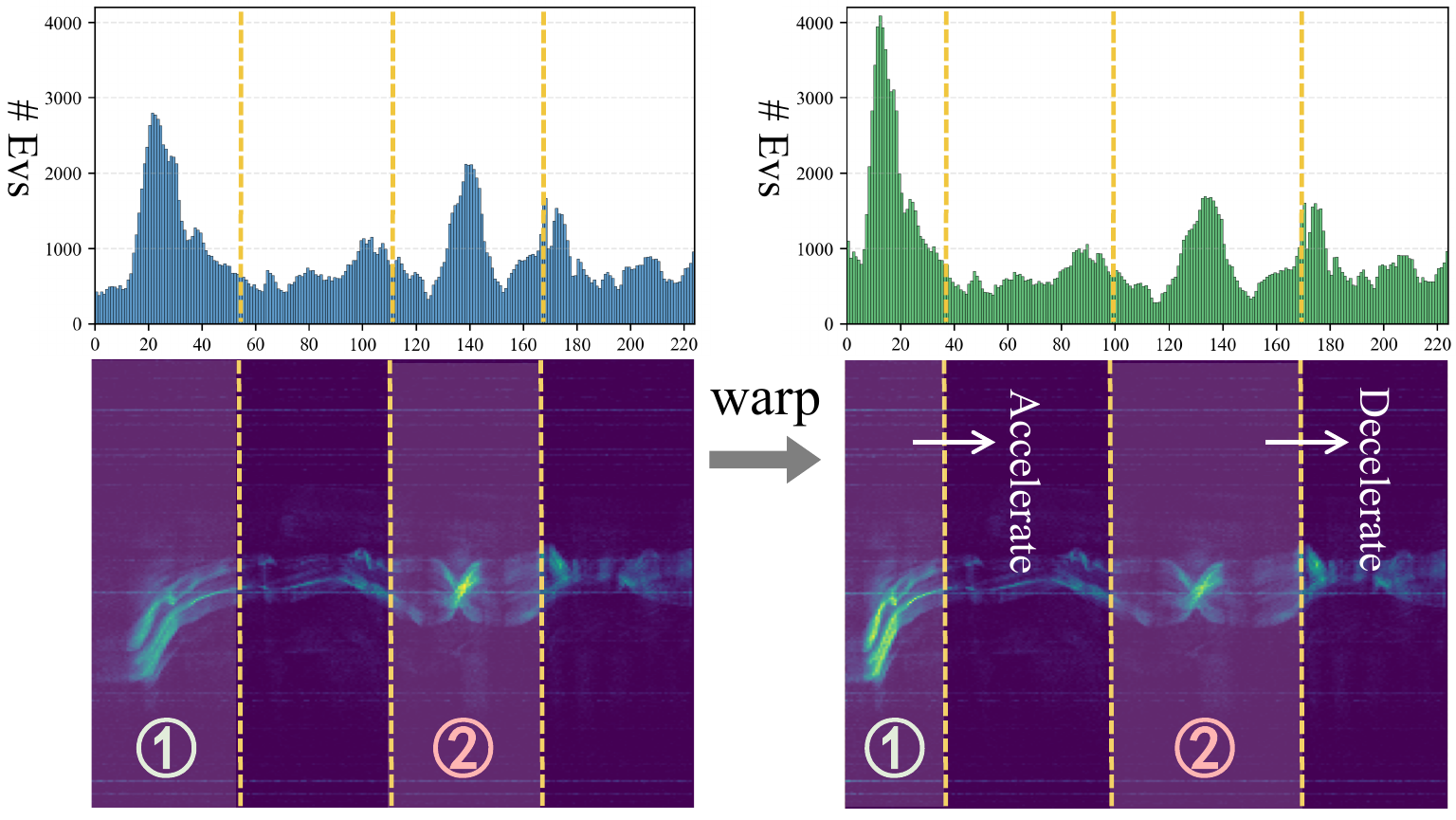}  
\caption{An example of DTW-warped $T$-$H$ map and the resulting event density change on 1-D event count along T dimension.}  
\label{fig:7}  
\end{figure}

\noindent\textbf{Implementation Details.} All experiments are trained for 80 epochs in PyTorch~\cite{paszke2019pytorch}. We use AdamW optimizer~\cite{loshchilov2017decoupled} with learning rate 1e-4 and weight decay 1e-5. Following~\cite{deng2021mvf}, $T$-axis are discretized into $\text{T}=224$ and $F_{th},F_{tw}$ are resized into $224\times224$. $l$ of DTW is set to 4 experimentally (Supplementary). Some results replicated from other methods are obtained following their original configurations through aligned training setup with our SVM-EAR* (without augmentation), details in the Supplementary. 

\begin{table}[!t]
\centering
\caption{Comparisons on THU-EACT-50-CHL~\cite{gao2023action} dataset.}
\label{tab:3}
\scriptsize
\setlength{\tabcolsep}{2pt}
\begin{tabular}{l|c|c|c|c}
\toprule 
\textbf{Method} & \textbf{Input Type} & \textbf{Top-1}(\%)$\uparrow$ & \textbf{FLOPs}$\downarrow$ & \textbf{Params}$\downarrow$ \\ 
\midrule 
HMAX SNN~\cite{xiao2019event} & \multirow{3}{*}{Spike} & 32.7(-) & - & - \\
Motion SNN~\cite{liu2021event} & & 47.3(-) & - & - \\
EventMamba~\cite{chen2024spikmamba} &  & 59.4(-) & 3.7G & 0.91M \\
\midrule 
EV-ACT~\cite{gao2023action} & \multirow{1}{*}{Frame} & 58.5(-) & 0.95G & 21.3M \\
\midrule 
EventMG~\cite{wueventmg} & \multirow{1}{*}{Graph} & 58.8(-) & 0.69G & 0.77M \\
\midrule 
CoST~\cite{li2019collaborative} & \multirow{5}{*}{Multi-View} & 51.7($\pm$0.32) & 19.2G & 25.4M \\
MVFNet~\cite{wu2021mvfnet} & & 59.6($\pm$0.61) & 32.8G & 23.6M \\
MVF-Net~\cite{deng2021mvf} & & 56.5($\pm$0.25) & 5.6G & 33.6M \\
\rowcolor{lightblue} SMV-EAR*(ours) &  & 62.9($\pm$0.31) & 3.6G & 23.5M \\
\rowcolor{lightblue} SMV-EAR(ours) &  & 66.7($\pm$0.29) & 3.6G & 23.5M \\
\bottomrule
\end{tabular}
\end{table}

\subsection{Comparison with the State-of-the-Arts}
\label{sec:4_2}
We compare with across different benchmarks~\cite{gao2023action,wang2024hardvs,wang2024dailydvs} and paradigms. As shown in Tab.~\ref{tab:2}, Tab.~\ref{tab:3}, the Spike-input frameworks (SDT, Spikeformer, HMAX SNN, Motion SNN) maintain sparsity but exhibit lower accuracy then others. The token-input frameworks (TimeSformer, V-SwinTrans) achieve higher accuracy due to Transformer's powerful modeling capabilities but often require significant parameter and computational cost. Our SMV-EAR consistently achieves SOTA accuracy across all datasets with +7.0\% on HARDVS, +10.7\% on DailyDVS-200, +10.2\% on THU-EACT-50-CHL compared to the baseline method MVF-Net, while strictly controlling parameters (23.5M) and computational overhead (1.8G MACs). This validates our argument that principled SMVRL representation (TISM), architecture (DDCF) and augmentation (DTW) are more effective than merely increasing model complexity or adapting existing video or event SMVRL methods (Sec.~\ref{sec:1}).

\subsection{Ablation Studies}
\label{sec:4_3}
\noindent\textbf{Contributions.} We conduct comprehensive ablations to validate our design choices. Tab.~\ref{tab:4} analyzes the complementarity of our contributions on THU-EACT-50-CHL (the same applies below). With MVF-Net~\cite{deng2021mvf} as baseline, TISM mechanism provides significant gain (+2.9\%), showing that translation-invariant representation is crucial. DDCF contributes +3.5\% by enabling principled dual-branch extraction and dynamic cross-view interaction, while DTW adds +3.8\% through realistic action variations. This showcases the complementarity of our three contributions.

\begin{table}[!t]
\centering
\caption{Effectiveness of three contributions in our SMV-EAR.}
\label{tab:4}
\scriptsize
\setlength{\tabcolsep}{5pt}
\begin{tabular}{l|c|c|c|c}
\toprule
\textbf{Components} & \textbf{Top-1(\%)}$\uparrow$ & \textbf{FLOPs}$\downarrow$ & \textbf{Params}$\downarrow$ & \textbf{T(all)}$\downarrow$ \\
\midrule 
MVF-Net(baseline) & 56.5($\pm$0.25) & 5.6G & 33.6M & 14.0ms \\
+ TISM(w/ MVF-Net) & 59.4($\pm$0.17) & 5.5G & 33.6M & 13.8ms \\
+ TISM,DDCF & 62.9($\pm$0.31) & 3.6G & 23.5M & 10.6ms \\
\rowcolor{lightblue} + All & 66.7($\pm$0.29) & 3.6G & 23.5M & 10.6ms \\
\bottomrule
\end{tabular}
\end{table}

\begin{table}[!t]
\centering
\scriptsize
\caption{Effectiveness of TISM. TI: translation-invariance.}
\label{tab:5}
\setlength{\tabcolsep}{4pt}
\begin{tabular}{lcc|c|c|c|c}
\toprule
\textbf{$w_c$} & \textbf{$m_c$}  & \textbf{$a_c$} & \textbf{TI}  & \textbf{\#C} & \textbf{Top-1(\%)} & \textbf{T(cpu)} \\ \midrule 
\rowcolor{lightgray} $w_0$ & $c$ & \textit{sum} & $\checkmark$ & 1 & 63.7($\pm$0.12) & 1.2ms \\ 
$w_1$,$w_2$ & $c$ & \textit{sum} & $\times$ & 2 & 57.6($\pm$0.80) & 2.3ms \\
\rowcolor{lightgray} $w_0$ & $p$ & \textit{sum} & $\checkmark$ & 1 & 62.2($\pm$0.70) & 1.3ms \\
$w_3$,$w_4$,$w_5$ & $p$ & \textit{sum} & $\times$ & 3 & 55.7($\pm$0.14) & 2.9ms \\ \midrule 
$w_0$ & $z$ & \textit{sum} & $\times$ & 1 & 46.1($\pm$0.95) & 1.6ms \\
$w_0$ & $z$ & \textit{max} & $\times$ & 1 & 47.6($\pm$0.58) & 1.8ms \\
$w_0$ & $z$ & \textit{mean} & $\times$ & 1 & 49.3($\pm$0.80) & 1.7ms \\
\rowcolor{lightgray} $w_0$ & $z$ & \textit{var.} & $\checkmark$ & 1 & 56.2($\pm$0.89) & 4.1ms \\ \midrule 
\rowcolor{lightgray} $w_0$ & $c_+,c_-$ & \textit{sum} & $\checkmark$ & 2 & 62.7($\pm$0.45) & 2.3ms\\
$w_0$ & $z_+,z_-$ & \textit{max}& $\times$ & 2 & 45.0($\pm$0.81) & 3.5ms \\
$w_0$ & $z$ & \textit{max,min} & $\times$ & 2 & 46.8($\pm$0.18) & 3.9ms \\ \midrule 
$w_0$ & $c,p,z$ & \textit{sum,var.} & $\checkmark$ & 3 & 67.1($\pm$0.69) & 6.8ms \\ 
\rowcolor{lightblue} $w_0$ & $c,p$ & \textit{sum} & $\checkmark$ & \textbf{2} & 66.7($\pm$0.29) & 2.5ms \\
\bottomrule
\end{tabular}
\end{table}

\noindent\textbf{TISM} Tab.~\ref{tab:5} validates the effectiveness of TISM (Sec.~\ref{sec:3_2}). The translation-invariant window, measurement and aggregation functions surpass those translation-variant. Moreover, while introducing $(w_0,z,\textit{variance})$ channel leads to +0.4\% accuracy, its events conversion is time-consuming so that fall short in efficiency. We therefore convert events with measurements $c,p$ and aggregation \textit{sum} for each view. We also evaluate under controlled spatial shifts in Tab.~\ref{tab:6}.\, where our TISM remains stable (slight degradation results from out-of-bound valid events) while MVF-Net shows clear performance drops as the shift magnitude increases. These results thoroughly validate our analysis in Sec.~\ref{sec:3_2}.  

\noindent\textbf{DDCF.} Tab.~\ref{tab:7} confirms the effectiveness of the proposed DDCF (Sec.~\ref{sec:3_3}). Adding $F_{hw}$ yields limited gains and additional branch cost than the $F_{th},F_{tw}$ setup, this confirms our insight that omit $F_{hw}$ map for efficient EAR. Our sample-wise weighting (SW) shows superiority over MVF-Net's early-concatenation (EC)~\cite{deng2021mvf} (+10.8\%), logit-averaging (LA) (+2.4\%), view-wise weighting (VW) (+1.9\%) and class-wise weighting (CW) (+1.5\%) fusion without significantly compromising efficiency. In Tab.~\ref{tab:8}, learning $w_{th},w_{tw}$ from $S$ exhibits optimal accuracy-efficiency trade-off, validating extraction timing discussed in Sec.~\ref{sec:3_3}. Implementation choices of $\mathcal{T}(\cdot)$ are ablated in Supplementary, showing 2$\times$512 input multi-head attention achieves better modeling for cross-view complementarity.

\noindent\textbf{DTW.} Tab.~\ref{tab:9} validates the effectiveness of DTW. Combination of warping functions improve the accuracy from 62.9\% up to 66.7\%. Other augmentation methods~\cite{gu2021eventdrop,shen2023eventmix,bendig2024shapeaug,dong2025eventzoom} are also replicated and compared, the results reveal clear advantages (+3.3\% on EventZoom~\cite{dong2025eventzoom}), validating the specificity of DTW to EAR. Notably, it can be used in conjunction with other augmentations for improvement. Beyond accuracy, Fig.~\ref{fig:7} quantifies DTW’s intended effect on temporal density, corroborating its speed-variation modeling.
\noindent\textbf{Others.} To examine the sensitivity and robustness, additional ablations are provided in Supplementary including: (i) \textit{Architecture choice of $\mathcal{T}(\cdot)$}: comparing different implementations of $\mathcal{T}(\cdot)$, confirming a multi-head attention layer as optimal choice; (ii) \textit{Backbone generalization}: evaluating different ResNet variants (ResNet-18, 34, 50) to verify SMV-EAR's compact design philosophy; (iii) \textit{Temporal resolution}: assessing performance under varying $\text{T}\in\{56,112,224,448\}$ to demonstrate scalability; (iv) \textit{Training efficiency}: comparing convergence curves and training time against baselines; (v) \textit{Computational breakdown}: analyzing each module's contribution to overall computation.

\begin{table}[t]  
\centering  
\caption{Translation-robustness under controlled shifts on $z$-axis.}  
\label{tab:6}  
\scriptsize  
\begin{tabular}{l|ccccc}  
\toprule  
Method & 0px & $\pm$10px & $\pm$20px & $\pm$30px & $\pm$40px \\
\midrule  
MVF-Net & 56.5 & 51.5 & 51.3 & 49.7 & 46.1 \\
\rowcolor{lightblue} SMV-EAR & 66.7 & 66.0 & 65.7 & 65.5 & 64.9 \\
\bottomrule  
\end{tabular}  
\end{table}  

\begin{table}[!t]
\centering
\scriptsize
\caption{Effectiveness of dual-brach and SW fusion in DDCF.}
\label{tab:7}
\setlength{\tabcolsep}{4pt}
\begin{tabular}{l|c|c|c|c|c|c}
\toprule
 \textbf{View map} & \textbf{Branch} & \textbf{Fusion} & \textbf{Top-1(\%)}  & \textbf{FLOPs} & \textbf{Params} & \textbf{T(all)} \\ \midrule  
$F_{hw}$ & \textit{Single} & $\times$ & 35.6($\pm$0.28) & 1.78G & 11.2M & 6.1ms \\  
$F_{tw}$ & \textit{Single} & $\times$ & 51.3($\pm$0.09) & 1.78G & 11.2M & 6.1ms \\ 
$F_{th}$ & \textit{Single} & $\times$ & 60.9($\pm$0.37) & 1.78G & 11.2M & 6.1ms \\ 
$F_{hw}$,$F_{tw}$ & \textit{Dual} & \textit{SW} & 48.3($\pm$0.81) & 3.57G & 23.5M & 10.6ms \\ 
$F_{hw}$,$F_{th}$ & \textit{Dual} & \textit{SW} & 61.5($\pm$0.06) & 3.57G & 23.5M & 10.6ms \\ 
\rowcolor{lightblue} $F_{th}$,$F_{tw}$ & \textit{Dual} & \textit{SW} & 66.7($\pm$0.29) & 3.57G & 23.5M & 10.6ms \\ 
All & \textit{Triple} & \textit{SW} & 67.0($\pm$0.34) & 5.35G & 34.7M & 15.4ms \\
 \midrule  
$F_{th}$,$F_{tw}$ & \textit{Single} & \textit{EC} & 55.9($\pm$0.56) & 1.86G & 11.2M & 6.6ms \\ 
$F_{th}$,$F_{tw}$ & \textit{Dual} & \textit{LA} & 64.3($\pm$0.24) & 3.57G & 22.4M & 10.2ms \\
$F_{th}$,$F_{tw}$ & \textit{Dual} & \textit{VW} & 64.8($\pm$0.07) & 3.57G & 22.4M & 10.2ms \\ 
$F_{th}$,$F_{tw}$ & \textit{Dual} & \textit{CW} & 65.2($\pm$0.17) & 3.57G & 22.4M & 10.2ms \\ 
\rowcolor{lightblue} $F_{th}$,$F_{tw}$ & \textit{Dual} & \textit{SW} & 66.7($\pm$0.29) & 3.57G & 23.5M & 10.6ms
\\ 
\bottomrule
\end{tabular}
\end{table}

\subsection{Visualization and Analysis.}
\label{sec:4_4}
\begin{figure*}[!t]
    \centering
    \includegraphics[width=1.0\linewidth]{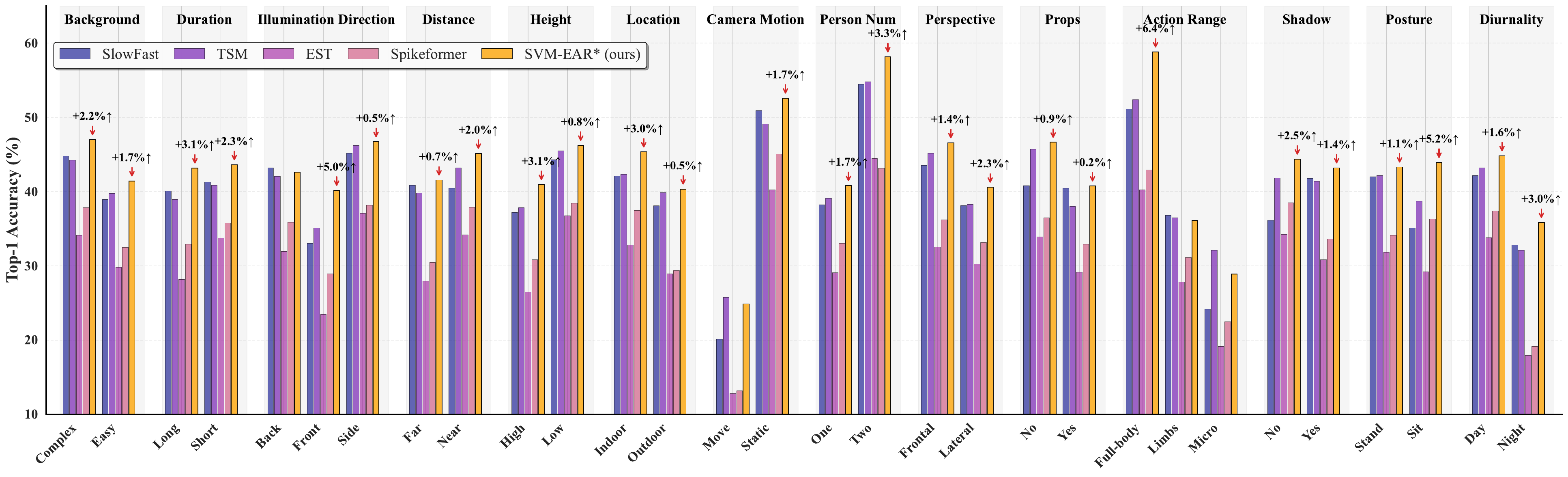}
    \caption{Comprehensive fine-grained evaluation and comparison across 14 different attribute categories on DailyDVS-200~\cite{wang2024dailydvs}.}
    \label{fig:8}
\end{figure*}

\begin{figure}[t]
    \centering
    \includegraphics[width=1.0\linewidth]{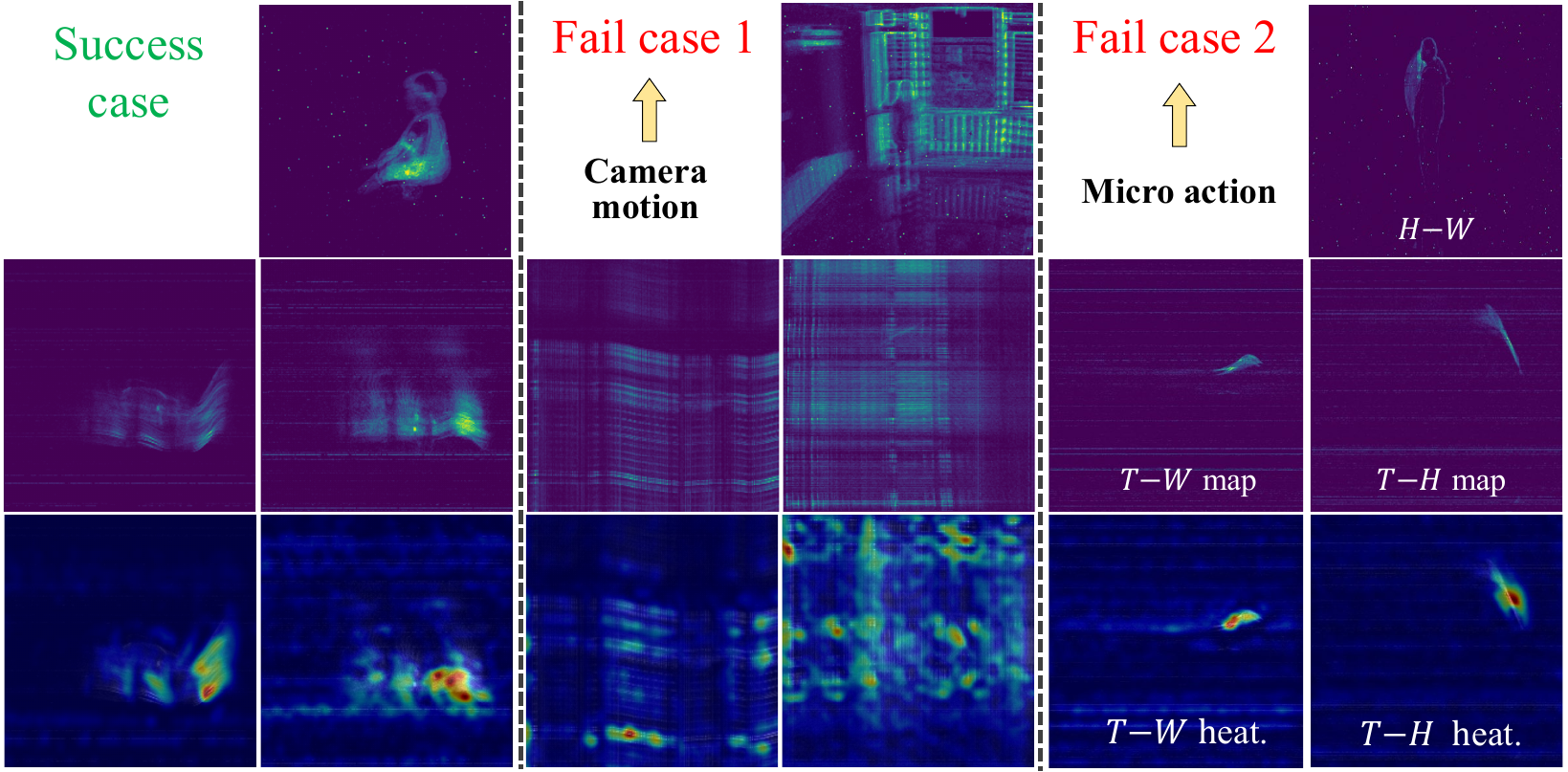}
    \caption{Heatmap visualizations. Successful cases highlight relevant regions, while failure cases show misdirected attention (in camera motion) or insufficient information (in micro action).}
    \label{fig:9}
\end{figure}

\noindent\textbf{Fine-grained Analysis.} Following prior work~\cite{wang2024dailydvs}, we conduct fine-grained analysis across different action attributes in Fig.~\ref{fig:8}. While SMV-EAR$^*$ achieves superior performance on most attributes, particularly for two-person interactions (+3.3\%) and sitting postures (+5.2\%), it also exposes specific limitations. Camera motion remains challenging (-2.9\% vs. SlowFast), as dense background events can occlude action-relevant patterns on temporal maps, degrading discriminability (Fig.~\ref{fig:9}). Interestingly, we observe that camera-induced motion exhibits near-linear streaks on temporal maps due to its globally uniform nature within short durations. This suggests a potential prior for decoupling foreground actions from background ego-motion through motion compensation or architectural inductive biases. Additionally, all methods exhibit degradation on micro actions (full-body vs. finger movements). This stems from limited spatiotemporal resolution, rendering micro-action regions informationally insufficient (Fig.~\ref{fig:9}). Potential remedies include higher-resolution sensors or adaptive attention mechanisms that selectively preserve fine-grained details. These analyses reveal principled directions for enhancing EAR under challenging real-world conditions.

\begin{table}[t]  
\centering  
\vspace{-0.2cm}
\caption{Extraction timing of $w_{th},w_{tw}$ (with additional Res-branches on concatenated temporal features then processed same as DDCF), $F^{(k)}$ denotes the feature after $k$-th Residual layer~\cite{he2016deep}.}
\label{tab:8}  
\scriptsize  
\begin{tabular}{l|c|c|c|c}  
\toprule  
\textbf{Temporal Feature} & \textbf{Top-1(\%)} & \textbf{FLOPs} & \textbf{Params} & \textbf{T(all)} \\
\midrule  
$[F_{th},F_{tw}]$ & 62.5($\pm$0.72) & 5.30G & 34.6M & 15.8ms \\
$[F^{(1)}_{th},F^{(1)}_{tw}]$ & 64.5($\pm$0.05) & 4.83G & 34.5M & 15.1ms \\
$[F^{(2)}_{th},F^{(2)}_{tw}]$ & 64.6($\pm$0.68) & 4.42G & 34.0M & 13.9ms \\
$[F^{(3)}_{th},F^{(3)}_{tw}]$ & 65.8($\pm$0.56) & 4.01G & 32.0M & 12.4ms \\
\rowcolor{lightblue} $[S_{th},S_{tw}]$ & 66.7($\pm$0.29) & 3.57G & 23.5M & 10.6ms \\
$[L_{th},L_{tw}]$ & 64.4($\pm$0.34) & 3.57G & 23.6M & 10.3ms \\
\bottomrule  
\end{tabular}  
\end{table}  

\begin{table}[!t]
\centering
\scriptsize
\caption{Effectiveness of DTW, $l=4$ intervals are applied, `-3D' denotes adapting original method to raw $H$-$W$-$T$ sparse events.}
\label{tab:9}
\setlength{\tabcolsep}{4pt}
\begin{tabular}{l|c|c}
\toprule 
 \textbf{Method} & \textbf{Warping} & \textbf{Top-1(\%)} \\ \midrule 
w/o Aug. & \textit{Id.} & 62.9($\pm$0.53) \\ 
DTW & +\textit{Lin.} & 64.2($\pm$0.76) \\ 
DTW & +\textit{Lin.,Pow.} & 65.3($\pm$0.23) \\ 
DTW & +\textit{Lin.,Pow.,Exp.} & 66.3($\pm$0.81) \\ 
\rowcolor{lightblue} DTW & +\textit{Lin.,Pow.,Exp.,Cos.} & 66.7($\pm$0.29) \\ \midrule  
EventDrop~\cite{gu2021eventdrop} & $\times$ & 62.7($\pm$0.40) \\ 
EventMix-3D~\cite{shen2023eventmix} & $\times$ & 63.1($\pm$0.96) \\ 
ShapeAug~\cite{bendig2024shapeaug} & $\times$ & 62.3($\pm$0.85) \\ 
EventZoom-3D~\cite{dong2025eventzoom} & $\times$ & 63.4($\pm$0.51) \\ 
\rowcolor{lightblue} DTW & +\textit{Lin.,Pow.,Exp.,Cos.} & 66.7($\pm$0.29) \\ 
\midrule
DTW+EventDrop & +\textit{Lin.,Pow.,Exp.,Cos.} & 67.6($\pm$0.17) \\ 
DTW+EventMix-3D & +\textit{Lin.,Pow.,Exp.,Cos.} & 67.4($\pm$0.09) \\ 
DTW+ShapeAug & +\textit{Lin.,Pow.,Exp.,Cos.} & 67.2($\pm$0.24) \\ 
DTW+EventZoom-3D & +\textit{Lin.,Pow.,Exp.,Cos.} & 68.1($\pm$0.74) \\ 
\bottomrule 
\end{tabular}
\end{table}

\noindent\textbf{Feature Visualization.} To provide intuitive understanding of SMV-EAR's learned representations, Fig.~\ref{fig:9} visualizes attention heatmaps overlaid on $T$-$H$ and $T$-$W$ maps. In successful cases, the model accurately localizes action-relevant regions, with complementary semantic cues emerging across temporal views, validating our dual-branch design. Conversely, failure cases exhibit attention diffusion under background clutter induced by camera motion or insufficient feature discriminability under micro action. These visualizations corroborate the quantitative findings in Fig.~\ref{fig:8}, offering interpretable insights into model behavior.

\begin{figure}[t]
    \centering
    \includegraphics[width=0.90\linewidth]{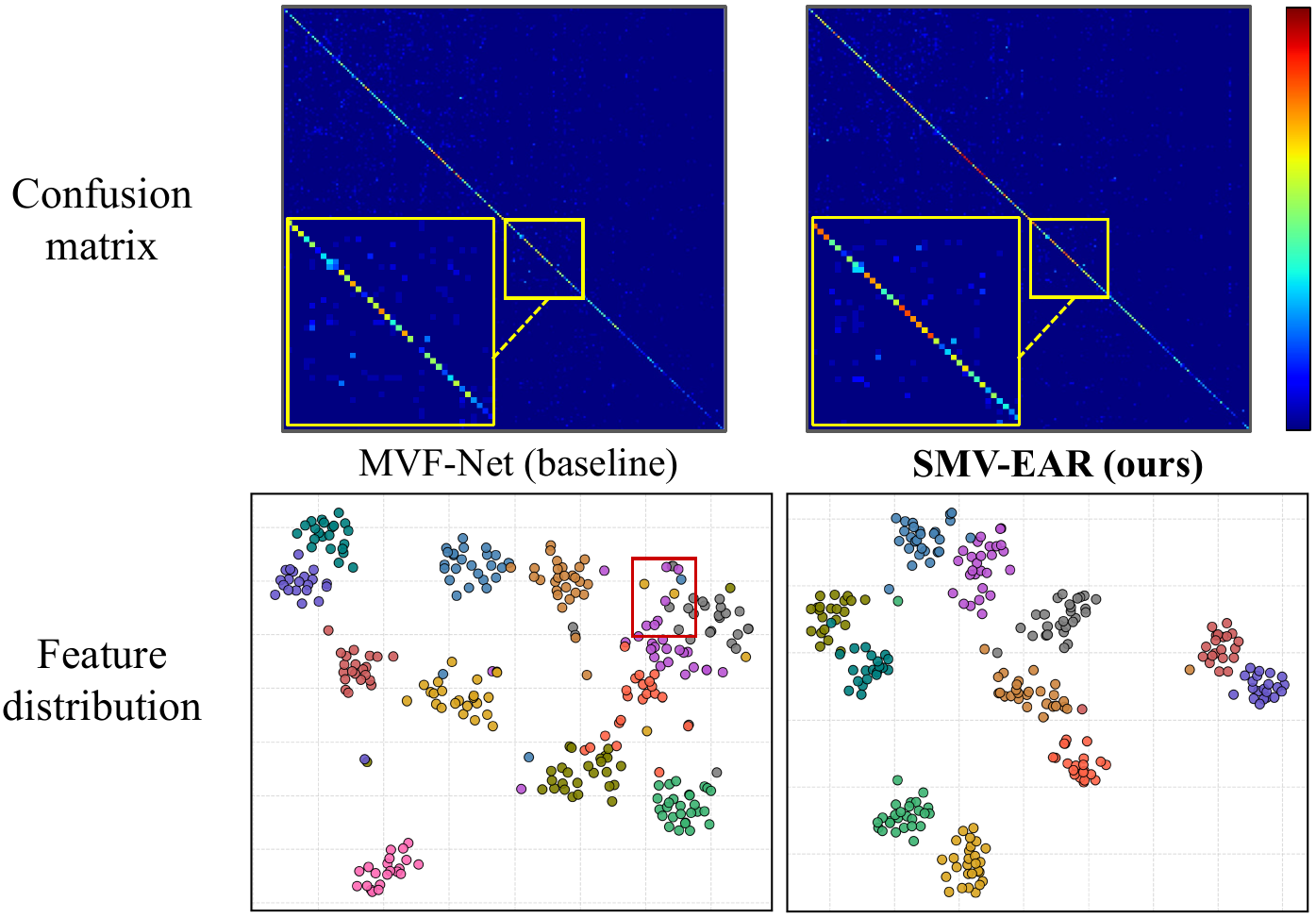}
    \caption{Visualization of confusion matrices (top) and feature distribution (bottom) of MVF-Net~\cite{deng2021mvf} and SMV-EAR. Ours shows improved cluster separation and reduced intra-class variance.}
    \label{fig:10}
\end{figure}

\noindent\textbf{Feature Discriminability.} Fig.~\ref{fig:10} illustrates feature distributions using t-SNE and presenting confusion matrices to compare our SMV-EAR with MVF-Net~\cite{deng2021mvf}. The t-SNE plot on action logits spanning 10 categories reveals that our method produces more distinct and well-separated clusters compared to MVF-Net, indicating superior feature discrimination. The confusion matrix on all classes further highlights the improvement of DDCF by exhibiting reduced confusion between different action categories.

\section{Limitation and Future Work}
\label{sec:5}
Despite superior performance, there are some limitations of our method deserve clarification to benefit the future EAR research. For representation, a background noise filter could be investigated to remove the linear, action-irrelevant background streaks on $T$-$H$ and $T$-$W$ maps (Fig.~\ref{fig:9}), thereby promising for improved discriminability of action representation. For architecture, an insightful observation that the motion trajectories on $T$-$H$ and $T$-$W$ maps exhibit non-stationary dynamics characterized by time-varying frequency components, hence a frequency domain processing architecture may be useful for more effective modeling of temporal motion dynamics. For augmentation, we notice that recent EventAugment~\cite{gu2024eventaugment} has explored learning-based strategies in $H$-$W$ view's event representation, the similar paradigm could be investigated for our TISM representation to unlock further generalization ability gains. We leave a full design and evaluation to the future work.

\section{Conclusion}
\label{sec:6}
This paper, we pioneer the first spatiotemporal multi-view representation learning (SMVRL) framework for EAR task. By revisiting key design stages spanning \textit{representation}, \textit{architecture} and \textit{augmentation}, we contribute a translation-invariant spatiotemporal multi-view event representation, a dual-branch cross-view dynamic interactive network and a fundamental temporal warping data augmentation, which enhance EAR performance on various challenging benchmarks effectively from different aspects, while with strictly controlled parameters and computational overhead. This establishes our principled SMV-EAR as a novel, powerful and efficient framework paradigm for EAR task with dense event representation. Beyond the scope of this article, we hope our SMVRL EAR approach will advance the understanding of event data's spatiotemporal dynamics, and ultimately benefit a wider spectrum of EAR-similar challenges in event-based vision in the future.

\bibliography{main}

\begin{thebibliography}{65}
\providecommand{\natexlab}[1]{#1}
\providecommand{\url}[1]{\texttt{#1}}
\expandafter\ifx\csname urlstyle\endcsname\relax
  \providecommand{\doi}[1]{doi: #1}\else
  \providecommand{\doi}{doi: \begingroup \urlstyle{rm}\Url}\fi

\bibitem[Bendig et~al.(2024)Bendig, Schuster, and Stricker]{bendig2024shapeaug}
Bendig, K., Schuster, R., and Stricker, D.
\newblock Shapeaug: Occlusion augmentation for event camera data.
\newblock \emph{arXiv preprint arXiv:2401.02274}, 2024.

\bibitem[Berahmand et~al.(2025)Berahmand, Daneshfar, Rahmaninia, Haghighat, and Jalili]{berahmand2025comprehensive}
Berahmand, K., Daneshfar, F., Rahmaninia, M., Haghighat, M., and Jalili, M.
\newblock A comprehensive survey on multi-view classification: Methods, applications, and challenges.
\newblock \emph{ACM Transactions on Intelligent Systems and Technology}, 2025.

\bibitem[Bertasius et~al.(2021)Bertasius, Wang, and Torresani]{bertasius2021space}
Bertasius, G., Wang, H., and Torresani, L.
\newblock Is space-time attention all you need for video understanding?
\newblock In \emph{Icml}, volume~2, pp.\ ~4, 2021.

\bibitem[Chakravarthi et~al.(2024)Chakravarthi, Verma, Daniilidis, Fermuller, and Yang]{chakravarthi2024recent}
Chakravarthi, B., Verma, A.~A., Daniilidis, K., Fermuller, C., and Yang, Y.
\newblock Recent event camera innovations: A survey.
\newblock In \emph{European Conference on Computer Vision}, pp.\  342--376. Springer, 2024.

\bibitem[Chen et~al.(2025)Chen, Huang, Wu, Liu, Bai, Shu, Yang, and Lim]{chen2025temporal}
Chen, H.~H., Huang, H., Wu, X., Liu, Y., Bai, Y., Shu, W.-J., Yang, H., and Lim, S.-N.
\newblock Temporal regularization makes your video generator stronger.
\newblock \emph{arXiv preprint arXiv:2503.15417}, 2025.

\bibitem[Chen et~al.(2024)Chen, Yang, Deng, Teng, and Pan]{chen2024spikmamba}
Chen, J., Yang, Y., Deng, S., Teng, D., and Pan, L.
\newblock Spikmamba: When snn meets mamba in event-based human action recognition.
\newblock In \emph{Proceedings of the 6th ACM International Conference on Multimedia in Asia}, pp.\  1--8, 2024.

\bibitem[de~Blegiers et~al.(2023)de~Blegiers, Dave, Yousaf, and Shah]{de2023eventtransact}
de~Blegiers, T., Dave, I.~R., Yousaf, A., and Shah, M.
\newblock Eventtransact: A video transformer-based framework for event-camera based action recognition.
\newblock In \emph{2023 IEEE/RSJ International Conference on Intelligent Robots and Systems (IROS)}, pp.\  1--7. IEEE, 2023.

\bibitem[Deng et~al.(2021)Deng, Chen, and Li]{deng2021mvf}
Deng, Y., Chen, H., and Li, Y.
\newblock Mvf-net: A multi-view fusion network for event-based object classification.
\newblock \emph{IEEE Transactions on Circuits and Systems for Video Technology}, 32\penalty0 (12):\penalty0 8275--8284, 2021.

\bibitem[Deng et~al.(2024)Deng, Chen, and Li]{deng2024dynamic}
Deng, Y., Chen, H., and Li, Y.
\newblock A dynamic gcn with cross-representation distillation for event-based learning.
\newblock In \emph{Proceedings of the AAAI Conference on Artificial Intelligence}, volume~38, pp.\  1492--1500, 2024.

\bibitem[Dong et~al.(2025)Dong, He, Shen, Zhao, Li, and Zeng]{dong2025eventzoom}
Dong, Y., He, X., Shen, G., Zhao, D., Li, Y., and Zeng, Y.
\newblock Eventzoom: A progressive approach to event-based data augmentation for enhanced neuromorphic vision.
\newblock In \emph{Proceedings of the AAAI Conference on Artificial Intelligence}, volume~39, pp.\  1291--1299, 2025.

\bibitem[Fan et~al.(2025)Fan, Hao, Guan, Rui, Gu, Wu, Zeng, and Zhu]{fan2025eventpillars}
Fan, R., Hao, W., Guan, J., Rui, L., Gu, L., Wu, T., Zeng, F., and Zhu, Z.
\newblock Eventpillars: Pillar-based efficient representations for event data.
\newblock In \emph{Proceedings of the AAAI Conference on Artificial Intelligence}, volume~39, pp.\  2861--2869, 2025.

\bibitem[Feichtenhofer et~al.(2019)Feichtenhofer, Fan, Malik, and He]{feichtenhofer2019slowfast}
Feichtenhofer, C., Fan, H., Malik, J., and He, K.
\newblock Slowfast networks for video recognition.
\newblock In \emph{Proceedings of the IEEE/CVF international conference on computer vision}, pp.\  6202--6211, 2019.

\bibitem[Fey et~al.(2018)Fey, Lenssen, Weichert, and M{\"u}ller]{fey2018splinecnn}
Fey, M., Lenssen, J.~E., Weichert, F., and M{\"u}ller, H.
\newblock Splinecnn: Fast geometric deep learning with continuous b-spline kernels.
\newblock In \emph{Proceedings of the IEEE conference on computer vision and pattern recognition}, pp.\  869--877, 2018.

\bibitem[Gallego et~al.(2020)Gallego, Delbr{\"u}ck, Orchard, Bartolozzi, Taba, Censi, Leutenegger, Davison, Conradt, Daniilidis, et~al.]{gallego2020event}
Gallego, G., Delbr{\"u}ck, T., Orchard, G., Bartolozzi, C., Taba, B., Censi, A., Leutenegger, S., Davison, A.~J., Conradt, J., Daniilidis, K., et~al.
\newblock Event-based vision: A survey.
\newblock \emph{IEEE transactions on pattern analysis and machine intelligence}, 44\penalty0 (1):\penalty0 154--180, 2020.

\bibitem[Gao et~al.(2023)Gao, Lu, Li, Ma, Du, Li, and Dai]{gao2023action}
Gao, Y., Lu, J., Li, S., Ma, N., Du, S., Li, Y., and Dai, Q.
\newblock Action recognition and benchmark using event cameras.
\newblock \emph{IEEE Transactions on Pattern Analysis and Machine Intelligence}, 45\penalty0 (12):\penalty0 14081--14097, 2023.

\bibitem[Gao et~al.(2024)Gao, Lu, Li, Li, and Du]{gao2024hypergraph}
Gao, Y., Lu, J., Li, S., Li, Y., and Du, S.
\newblock Hypergraph-based multi-view action recognition using event cameras.
\newblock \emph{IEEE Transactions on Pattern Analysis and Machine Intelligence}, 46\penalty0 (10):\penalty0 6610--6622, 2024.

\bibitem[Gehrig et~al.(2019)Gehrig, Loquercio, Derpanis, and Scaramuzza]{gehrig2019end}
Gehrig, D., Loquercio, A., Derpanis, K.~G., and Scaramuzza, D.
\newblock End-to-end learning of representations for asynchronous event-based data.
\newblock In \emph{Proceedings of the IEEE/CVF international conference on computer vision}, pp.\  5633--5643, 2019.

\bibitem[Gu et~al.(2021)Gu, Sng, Hu, and Yu]{gu2021eventdrop}
Gu, F., Sng, W., Hu, X., and Yu, F.
\newblock Eventdrop: Data augmentation for event-based learning.
\newblock \emph{arXiv preprint arXiv:2106.05836}, 2021.

\bibitem[Gu et~al.(2024)Gu, Dou, Li, Long, Guo, Chen, Liu, Jiao, and Li]{gu2024eventaugment}
Gu, F., Dou, J., Li, M., Long, X., Guo, S., Chen, C., Liu, K., Jiao, X., and Li, R.
\newblock Eventaugment: learning augmentation policies from asynchronous event-based data.
\newblock \emph{IEEE Transactions on Cognitive and Developmental Systems}, 16\penalty0 (4):\penalty0 1521--1532, 2024.

\bibitem[Hamann et~al.(2025)Hamann, Ghosh, Ju{\'a}rez~Mart{\'\i}nez, Hart, Kacelnik, and Gallego]{hamann2025fourier}
Hamann, F., Ghosh, S., Ju{\'a}rez~Mart{\'\i}nez, I., Hart, T., Kacelnik, A., and Gallego, G.
\newblock Fourier-based action recognition for wildlife behavior quantification with event cameras.
\newblock \emph{Advanced Intelligent Systems}, 7\penalty0 (2):\penalty0 2400353, 2025.

\bibitem[He et~al.(2016)He, Zhang, Ren, and Sun]{he2016deep}
He, K., Zhang, X., Ren, S., and Sun, J.
\newblock Deep residual learning for image recognition.
\newblock In \emph{Proceedings of the IEEE conference on computer vision and pattern recognition}, pp.\  770--778, 2016.

\bibitem[Li et~al.(2019)Li, Zhong, Xie, and Pu]{li2019collaborative}
Li, C., Zhong, Q., Xie, D., and Pu, S.
\newblock Collaborative spatiotemporal feature learning for video action recognition.
\newblock In \emph{Proceedings of the ieee/cvf conference on computer vision and pattern recognition}, pp.\  7872--7881, 2019.

\bibitem[Li et~al.(2022)Li, Kim, Park, Geller, and Panda]{li2022neuromorphic}
Li, Y., Kim, Y., Park, H., Geller, T., and Panda, P.
\newblock Neuromorphic data augmentation for training spiking neural networks.
\newblock In \emph{European Conference on Computer Vision}, pp.\  631--649. Springer, 2022.

\bibitem[Lin et~al.(2019)Lin, Gan, and Han]{lin2019tsm}
Lin, J., Gan, C., and Han, S.
\newblock Tsm: Temporal shift module for efficient video understanding.
\newblock In \emph{Proceedings of the IEEE/CVF international conference on computer vision}, pp.\  7083--7093, 2019.

\bibitem[Lin et~al.(2024)Lin, Liu, and Chen]{lin2024spike}
Lin, X., Liu, M., and Chen, H.
\newblock Spike-har++: an energy-efficient and lightweight parallel spiking transformer for event-based human action recognition.
\newblock \emph{Frontiers in Computational Neuroscience}, 18:\penalty0 1508297, 2024.

\bibitem[Liu et~al.(2021)Liu, Xing, Tang, Ma, and Pan]{liu2021event}
Liu, Q., Xing, D., Tang, H., Ma, D., and Pan, G.
\newblock Event-based action recognition using motion information and spiking neural networks.
\newblock In \emph{IJCAI}, pp.\  1743--1749, 2021.

\bibitem[Liu et~al.(2022)Liu, Ning, Cao, Wei, Zhang, Lin, and Hu]{liu2022video}
Liu, Z., Ning, J., Cao, Y., Wei, Y., Zhang, Z., Lin, S., and Hu, H.
\newblock Video swin transformer.
\newblock In \emph{Proceedings of the IEEE/CVF conference on computer vision and pattern recognition}, pp.\  3202--3211, 2022.

\bibitem[Loshchilov \& Hutter(2017)Loshchilov and Hutter]{loshchilov2017decoupled}
Loshchilov, I. and Hutter, F.
\newblock Decoupled weight decay regularization.
\newblock \emph{arXiv preprint arXiv:1711.05101}, 2017.

\bibitem[Miao et~al.(2019)Miao, Chen, Ning, Zi, Ren, Bing, and Knoll]{miao2019neuromorphic}
Miao, S., Chen, G., Ning, X., Zi, Y., Ren, K., Bing, Z., and Knoll, A.
\newblock Neuromorphic vision datasets for pedestrian detection, action recognition, and fall detection.
\newblock \emph{Frontiers in neurorobotics}, 13:\penalty0 38, 2019.

\bibitem[Neimark et~al.(2021)Neimark, Bar, Zohar, and Asselmann]{neimark2021video}
Neimark, D., Bar, O., Zohar, M., and Asselmann, D.
\newblock Video transformer network.
\newblock In \emph{Proceedings of the IEEE/CVF international conference on computer vision}, pp.\  3163--3172, 2021.

\bibitem[Paszke et~al.(2019)Paszke, Gross, Massa, Lerer, Bradbury, Chanan, Killeen, Lin, Gimelshein, Antiga, et~al.]{paszke2019pytorch}
Paszke, A., Gross, S., Massa, F., Lerer, A., Bradbury, J., Chanan, G., Killeen, T., Lin, Z., Gimelshein, N., Antiga, L., et~al.
\newblock Pytorch: An imperative style, high-performance deep learning library.
\newblock \emph{Advances in neural information processing systems}, 32, 2019.

\bibitem[Peng et~al.(2023)Peng, Zhang, Xiong, Sun, and Wu]{peng2023get}
Peng, Y., Zhang, Y., Xiong, Z., Sun, X., and Wu, F.
\newblock Get: Group event transformer for event-based vision.
\newblock In \emph{Proceedings of the IEEE/CVF International Conference on Computer Vision}, pp.\  6038--6048, 2023.

\bibitem[Qi et~al.(2017{\natexlab{a}})Qi, Su, Mo, and Guibas]{qi2017pointnet}
Qi, C.~R., Su, H., Mo, K., and Guibas, L.~J.
\newblock Pointnet: Deep learning on point sets for 3d classification and segmentation.
\newblock In \emph{Proceedings of the IEEE conference on computer vision and pattern recognition}, pp.\  652--660, 2017{\natexlab{a}}.

\bibitem[Qi et~al.(2017{\natexlab{b}})Qi, Yi, Su, and Guibas]{qi2017pointnet++}
Qi, C.~R., Yi, L., Su, H., and Guibas, L.~J.
\newblock Pointnet++: Deep hierarchical feature learning on point sets in a metric space.
\newblock \emph{Advances in neural information processing systems}, 30, 2017{\natexlab{b}}.

\bibitem[Ramesh et~al.(2023)Ramesh, Dall’Alba, Gonzalez, Yu, Mascagni, Mutter, Marescaux, Fiorini, and Padoy]{ramesh2023trandaugment}
Ramesh, S., Dall’Alba, D., Gonzalez, C., Yu, T., Mascagni, P., Mutter, D., Marescaux, J., Fiorini, P., and Padoy, N.
\newblock Trandaugment: temporal random augmentation strategy for surgical activity recognition from videos.
\newblock \emph{International Journal of Computer Assisted Radiology and Surgery}, 18\penalty0 (9):\penalty0 1665--1672, 2023.

\bibitem[Rebecq et~al.(2017)Rebecq, Horstschaefer, and Scaramuzza]{rebecq2017real}
Rebecq, H., Horstschaefer, T., and Scaramuzza, D.
\newblock Real-time visual-inertial odometry for event cameras using keyframe-based nonlinear optimization.
\newblock \emph{British Machine Vision Conference}, 2017.

\bibitem[Ren et~al.(2023)Ren, Zhou, Huang, Fu, Lin, Song, and Cheng]{ren2023spikepoint}
Ren, H., Zhou, Y., Huang, Y., Fu, H., Lin, X., Song, J., and Cheng, B.
\newblock Spikepoint: An efficient point-based spiking neural network for event cameras action recognition.
\newblock \emph{arXiv preprint arXiv:2310.07189}, 2023.

\bibitem[Ruan et~al.(2025)Ruan, Pu, Chen, Gao, Guo, Kong, Xie, and Wei]{ruan2025few}
Ruan, Z., Pu, N., Chen, J., Gao, S., Guo, Y., Kong, Q., Xie, Y., and Wei, Y.
\newblock Few-shot event-based action recognition.
\newblock \emph{Neural Networks}, pp.\  107750, 2025.

\bibitem[Schaefer et~al.(2022)Schaefer, Gehrig, and Scaramuzza]{schaefer2022aegnn}
Schaefer, S., Gehrig, D., and Scaramuzza, D.
\newblock Aegnn: Asynchronous event-based graph neural networks.
\newblock In \emph{Proceedings of the IEEE/CVF conference on computer vision and pattern recognition}, pp.\  12371--12381, 2022.

\bibitem[Shen et~al.(2023)Shen, Zhao, and Zeng]{shen2023eventmix}
Shen, G., Zhao, D., and Zeng, Y.
\newblock Eventmix: An efficient data augmentation strategy for event-based learning.
\newblock \emph{Information Sciences}, 644:\penalty0 119170, 2023.

\bibitem[Steffen et~al.(2024)Steffen, Trapp, Roennau, and Dillmann]{steffen2024efficient}
Steffen, L., Trapp, T., Roennau, A., and Dillmann, R.
\newblock Efficient gesture recognition on spiking convolutional networks through sensor fusion of event-based and depth data.
\newblock In \emph{2024 IEEE International Conference on Robotics and Automation (ICRA)}, pp.\  345--352. IEEE, 2024.

\bibitem[Sun et~al.(2025)Sun, Zhang, Wang, Cao, and Xu]{sun2025event}
Sun, J., Zhang, Q., Wang, J., Cao, J., and Xu, R.
\newblock Event masked autoencoder: Point-wise action recognition with event-based cameras.
\newblock \emph{arXiv preprint arXiv:2501.01040}, 2025.

\bibitem[Sun et~al.(2023)Sun, Zhang, Cheng, and Lu]{sun2023asynchronous}
Sun, L., Zhang, Y., Cheng, J., and Lu, H.
\newblock Asynchronous event processing with local-shift graph convolutional network.
\newblock In \emph{Proceedings of the AAAI Conference on Artificial Intelligence}, volume~37, pp.\  2402--2410, 2023.

\bibitem[Tavanaei et~al.(2019)Tavanaei, Ghodrati, Kheradpisheh, Masquelier, and Maida]{tavanaei2019deep}
Tavanaei, A., Ghodrati, M., Kheradpisheh, S.~R., Masquelier, T., and Maida, A.
\newblock Deep learning in spiking neural networks.
\newblock \emph{Neural networks}, 111:\penalty0 47--63, 2019.

\bibitem[Tran et~al.(2015)Tran, Bourdev, Fergus, Torresani, and Paluri]{tran2015learning}
Tran, D., Bourdev, L., Fergus, R., Torresani, L., and Paluri, M.
\newblock Learning spatiotemporal features with 3d convolutional networks.
\newblock In \emph{Proceedings of the IEEE international conference on computer vision}, pp.\  4489--4497, 2015.

\bibitem[Tran et~al.(2018)Tran, Wang, Torresani, Ray, LeCun, and Paluri]{r2plus1d}
Tran, D., Wang, H., Torresani, L., Ray, J., LeCun, Y., and Paluri, M.
\newblock A closer look at spatiotemporal convolutions for action recognition.
\newblock In \emph{Proceedings of the IEEE conference on Computer Vision and Pattern Recognition}, pp.\  6450--6459, 2018.

\bibitem[Vaswani et~al.(2017)Vaswani, Shazeer, Parmar, Uszkoreit, Jones, Gomez, Kaiser, and Polosukhin]{vaswani2017attention}
Vaswani, A., Shazeer, N., Parmar, N., Uszkoreit, J., Jones, L., Gomez, A.~N., Kaiser, {\L}., and Polosukhin, I.
\newblock Attention is all you need.
\newblock \emph{Advances in neural information processing systems}, 30, 2017.

\bibitem[Wang et~al.(2024{\natexlab{a}})Wang, Xu, Lin, Ye, Li, Zhu, Ali~Shah, Bennamoun, and Zhang]{wang2024dailydvs}
Wang, Q., Xu, Z., Lin, Y., Ye, J., Li, H., Zhu, G., Ali~Shah, S.~A., Bennamoun, M., and Zhang, L.
\newblock Dailydvs-200: A comprehensive benchmark dataset for event-based action recognition.
\newblock In \emph{European Conference on Computer Vision}, pp.\  55--72. Springer, 2024{\natexlab{a}}.

\bibitem[Wang et~al.(2024{\natexlab{b}})Wang, Wu, Jiang, Bao, Zhu, Li, Wang, and Tian]{wang2024hardvs}
Wang, X., Wu, Z., Jiang, B., Bao, Z., Zhu, L., Li, G., Wang, Y., and Tian, Y.
\newblock Hardvs: Revisiting human activity recognition with dynamic vision sensors.
\newblock In \emph{Proceedings of the AAAI conference on artificial intelligence}, volume~38, pp.\  5615--5623, 2024{\natexlab{b}}.

\bibitem[Wang et~al.(2025)Wang, Wang, Wang, Chen, Jin, Song, Jiang, and Li]{wang2025rgb}
Wang, X., Wang, H., Wang, S., Chen, Q., Jin, J., Song, H., Jiang, B., and Li, C.
\newblock Rgb-event based pedestrian attribute recognition: A benchmark dataset and an asymmetric rwkv fusion framework.
\newblock \emph{arXiv preprint arXiv:2504.10018}, 2025.

\bibitem[Wang et~al.(2019)Wang, Sun, Liu, Sarma, Bronstein, and Solomon]{wang2019dynamic}
Wang, Y., Sun, Y., Liu, Z., Sarma, S.~E., Bronstein, M.~M., and Solomon, J.~M.
\newblock Dynamic graph cnn for learning on point clouds.
\newblock \emph{ACM Transactions on Graphics (tog)}, 38\penalty0 (5):\penalty0 1--12, 2019.

\bibitem[Wang et~al.(2021)Wang, She, and Smolic]{wang2021action}
Wang, Z., She, Q., and Smolic, A.
\newblock Action-net: Multipath excitation for action recognition.
\newblock In \emph{Proceedings of the IEEE/CVF conference on computer vision and pattern recognition}, pp.\  13214--13223, 2021.

\bibitem[Wu et~al.(2025)Wu, Ma, and Li]{wu2025transformer}
Wu, H., Ma, X., and Li, Y.
\newblock Transformer-based multiview spatiotemporal feature interactive fusion for human action recognition in depth videos.
\newblock \emph{Signal Processing: Image Communication}, 131:\penalty0 117244, 2025.

\bibitem[Wu et~al.()Wu, Jin, Feng, and Hu]{wueventmg}
Wu, S., Jin, L., Feng, H., and Hu, B.
\newblock Eventmg: Efficient multilevel mamba-graph learning for spatiotemporal event representation.
\newblock In \emph{The Thirty-ninth Annual Conference on Neural Information Processing Systems}.

\bibitem[Wu et~al.(2024)Wu, Sheng, Feng, and Hu]{wu2024egsst}
Wu, S., Sheng, H., Feng, H., and Hu, B.
\newblock Egsst: Event-based graph spatiotemporal sensitive transformer for object detection.
\newblock \emph{Advances in Neural Information Processing Systems}, 37:\penalty0 120526--120548, 2024.

\bibitem[Wu et~al.(2021)Wu, He, Lin, Li, Gan, and Ding]{wu2021mvfnet}
Wu, W., He, D., Lin, T., Li, F., Gan, C., and Ding, E.
\newblock Mvfnet: Multi-view fusion network for efficient video recognition.
\newblock In \emph{Proceedings of the AAAI conference on artificial intelligence}, volume~35, pp.\  2943--2951, 2021.

\bibitem[Wu et~al.(2018)Wu, Deng, Li, Zhu, and Shi]{wu2018spatio}
Wu, Y., Deng, L., Li, G., Zhu, J., and Shi, L.
\newblock Spatio-temporal backpropagation for training high-performance spiking neural networks.
\newblock \emph{Frontiers in neuroscience}, 12:\penalty0 331, 2018.

\bibitem[Xiao et~al.(2019)Xiao, Tang, Ma, Yan, and Orchard]{xiao2019event}
Xiao, R., Tang, H., Ma, Y., Yan, R., and Orchard, G.
\newblock An event-driven categorization model for aer image sensors using multispike encoding and learning.
\newblock \emph{IEEE transactions on neural networks and learning systems}, 31\penalty0 (9):\penalty0 3649--3657, 2019.

\bibitem[Xie et~al.(2022)Xie, Deng, Shao, Liu, and Li]{xie2022vmv}
Xie, B., Deng, Y., Shao, Z., Liu, H., and Li, Y.
\newblock Vmv-gcn: Volumetric multi-view based graph cnn for event stream classification.
\newblock \emph{IEEE Robotics and Automation Letters}, 7\penalty0 (2):\penalty0 1976--1983, 2022.

\bibitem[Xie et~al.(2024)Xie, Deng, Shao, Xu, and Li]{xie2024event}
Xie, B., Deng, Y., Shao, Z., Xu, Q., and Li, Y.
\newblock Event voxel set transformer for spatiotemporal representation learning on event streams.
\newblock \emph{IEEE Transactions on Circuits and Systems for Video Technology}, 2024.

\bibitem[Yao et~al.(2023)Yao, Hu, Zhou, Yuan, Tian, Xu, and Li]{yao2023spike}
Yao, M., Hu, J., Zhou, Z., Yuan, L., Tian, Y., Xu, B., and Li, G.
\newblock Spike-driven transformer.
\newblock \emph{Advances in neural information processing systems}, 36:\penalty0 64043--64058, 2023.

\bibitem[Zhou et~al.(2024)Zhou, Zheng, Lyu, and Wang]{zhou2024exact}
Zhou, J., Zheng, X., Lyu, Y., and Wang, L.
\newblock Exact: Language-guided conceptual reasoning and uncertainty estimation for event-based action recognition and more.
\newblock In \emph{Proceedings of the IEEE/CVF Conference on Computer Vision and Pattern Recognition}, pp.\  18633--18643, 2024.

\bibitem[Zhou et~al.(2022)Zhou, Zhu, He, Wang, Yan, Tian, and Yuan]{zhou2022spikformer}
Zhou, Z., Zhu, Y., He, C., Wang, Y., Yan, S., Tian, Y., and Yuan, L.
\newblock Spikformer: When spiking neural network meets transformer.
\newblock \emph{arXiv preprint arXiv:2209.15425}, 2022.

\bibitem[Zhu et~al.(2019)Zhu, Yuan, Chaney, and Daniilidis]{zhu2019unsupervised}
Zhu, A.~Z., Yuan, L., Chaney, K., and Daniilidis, K.
\newblock Unsupervised event-based learning of optical flow, depth, and egomotion.
\newblock In \emph{Proceedings of the IEEE/CVF conference on computer vision and pattern recognition}, pp.\  989--997, 2019.

\bibitem[Zubi{\'c} et~al.(2023)Zubi{\'c}, Gehrig, Gehrig, and Scaramuzza]{zubic2023chaos}
Zubi{\'c}, N., Gehrig, D., Gehrig, M., and Scaramuzza, D.
\newblock From chaos comes order: Ordering event representations for object recognition and detection.
\newblock In \emph{Proceedings of the IEEE/CVF International Conference on Computer Vision}, pp.\  12846--12856, 2023.

\end{thebibliography}
\bibliographystyle{icml2025}




\end{document}